\documentclass{article}
\usepackage[utf8]{inputenc}
\usepackage{multicol} 
\usepackage[left=3cm, right=2.5cm, top=1.5cm, bottom=2.5cm]{geometry}
\usepackage{wrapfig} 
\usepackage{graphicx}
\usepackage{subfigure} 
\usepackage{multirow}
\usepackage{amsmath}
\usepackage{booktabs}
\usepackage{amssymb}
\usepackage{hyperref}
\usepackage{etoolbox} 
\usepackage{yax}      
\usepackage{float}
\usepackage[para,online,flushleft]{threeparttable}
\usepackage{authblk} 
\usepackage[numbers]{natbib} 
\usepackage{rotating}
\usepackage{array}
\usepackage{longtable}
\usepackage{threeparttable}
\usepackage{lscape}
\usepackage{pdflscape}

\hypersetup{
    colorlinks=true,
    linkcolor=blue,
    filecolor=blue,
    urlcolor=blue,
    citecolor=cyan,
}

\providecommand{\keywords}[1]
{
  \small	
  \textbf{\textit{Keywords:}} #1
}

\title{StatLLaMA: Multi-Stage training for domain-optimized statistical large language models}
\author[1]{Jing-Yi Zeng}
\author[,1]{Guan-Hua Huang\thanks{Corresponding author: ghuang@nycu.edu.tw}}
\affil[1]{Institute of Statistics, National Yang Ming Chiao Tung University, Hsinchu, Taiwan}
\date{}

\begin{document}

\maketitle

\bigskip

\begin{abstract}
\noindent This study investigates how to efficiently build a domain-specialized large language model (LLM) for statistics using the lightweight LLaMA-3.2-3B family as the foundation model (FM). We systematically compare three multi-stage training pipelines—starting from a base FM with no instruction-following capability, a base FM augmented with post-hoc instruction tuning, and an instruction-tuned FM with strong general reasoning abilities—across continual pretraining, supervised fine-tuning (SFT), reinforcement learning from human feedback (RLHF) preference alignment, and downstream task fine-tuning (DTFT). Results show that pipelines beginning with a base FM fail to develop meaningful statistical reasoning, even after extensive instruction tuning, SFT, or RLHF alignment. In contrast, starting from LLaMA-3.2-3B-Instruct enables effective domain specialization. A comprehensive evaluation of SFT variants reveals clear trade-offs between domain expertise and general reasoning ability. We further demonstrate that direct preference optimization provides stable and effective RLHF preference alignment. Finally, we show that DTFT must be performed with extremely low intensity to avoid catastrophic forgetting in highly optimized models. The final model, StatLLaMA, achieves strong and balanced performance on benchmarks of mathematical reasoning, common-sense reasoning, and statistical expertise, offering a practical blueprint for developing resource-efficient statistical LLMs. The code is available at \url{https://github.com/HuangDLab/StatLLaMA}.
\end{abstract}


\section{Introduction}

The development of large language models (LLMs) has profoundly reshaped natural language processing (NLP). Transformer-based architectures, such as BERT \citep{Devlinetal2019bertpre-training}, the GPT series \citep{Radfordetal2018improvinglanguage, Radfordetal2019languagemodels, Brownetal2020languagemodels}, and the recent LLaMA series \citep{Touvronetal2023llamaopen, Touvronetal2023llama2open, Grattafiorietal2024thellama3}, have demonstrated exceptional performance across a wide spectrum of general-purpose NLP tasks. These models have shifted the paradigm from zero-shot \citep{Weietal2022finetunedlanguage} and few-shot \citep{Brownetal2020languagemodels} learning toward more sophisticated instruction-following capabilities \citep{Ouyangetal2022traininglanguage}. Built upon large-scale pretraining using massive text corpora, these foundation models \citep{Bommasanietal2022onthe} encapsulate extensive world knowledge and general language understanding.

However, despite their impressive capabilities, general-purpose LLMs face substantial challenges when applied to specialized domains like statistics. These tasks require not only a deep understanding of domain-specific terminology but also the ability to perform multi-step reasoning and adhere to formal standards. General LLMs often exhibit limitations in these areas, resulting in inadequate performance and potentially misleading outputs. This raises a key research question: How can specialized domain knowledge be effectively integrated into general-purpose LLMs without compromising their general reasoning abilities? This study addresses this question by focusing on the statistics domain and designing a resource-efficient optimization and training pipeline for the lightweight LLaMA-3.2-3B model. Our objective is to develop a high-performance statistical LLM that retains general reasoning proficiency while gaining domain-specific competence.

Supervised fine-tuning (SFT) adapts general-purpose models to specific domains by training on input-output mapping pairs of domain tasks. Traditional implementations of SFT, such as used in BERT, often involve full fine-tuning (FFT), which updates all model parameters. While effective, FFT becomes prohibitively expensive as model sizes grow, making it unsuitable for deploying separate fine-tuned models per downstream task. Consequently, the research community has explored more efficient adaptation strategies. Early approaches like ULMFiT \citep{Howardetal2018universallanguage} introduced concepts such as discriminative fine-tuning and gradual unfreezing, marking the first steps toward parameter-efficient fine-tuning (PEFT). Recent advancements in PEFT, such as adapter tuning \citep{Houlsbyetal2019parameter-efficienttransfer}, prefix-tuning \citep{Lietal2021prefix-tuningoptimizing}, prompt tuning \citep{Lesteretal2021thepower}, and particularly low-rank adaptation (LoRA) \citep{Huetal2021loralow-rank}, enable efficient domain adaptation by freezing the majority of model parameters while training only a small number of task-specific parameters. These techniques reduce computational demands and often outperform in-context learning \citep{Liuetal2022few-shotparameter-efficient}. Tools like QLoRA \citep{Dettmersetal2023qloraefficient} further enhance efficiency by integrating quantization strategies. Nevertheless, the key challenge remains: how to inject domain knowledge deeply while preserving general reasoning. This study investigates this central question using PEFT as a foundational component.

Continual pretraining (CoP) \citep{Gururanganetal2020dontstop, Guptaetal2023continualpre-training, Jinetal2022lifelongpretraining} presents another avenue for domain adaptation, allowing the model to continue learning from domain-specific, unlabeled text. While effective for capturing linguistic and structural characteristics of the target domain, our preliminary experiments revealed that CoP alone without strong instruction-following capabilities yields limited improvements in complex reasoning tasks. Instruction tuning has emerged as a powerful technique for enhancing LLM usability and task generalization. By training on large collections of instruction-task-response triplets, models learn to better interpret and execute human instructions. From InstructGPT \citep{Ouyangetal2022traininglanguage} to FLAN \citep{Weietal2022finetunedlanguage, Chungetal2024scalinginstruction-finetuned, Longpreetal2023theflan} and Alpaca \citep{Taorietal2023alpacaa}, instruction-tuned models, often referred to as ``chat models'', have become the standard for interactive AI systems.

Yet, even SFT or CoP cannot fully ensure that a model aligns with human preferences in tasks involving intricate reasoning or subtle judgment. Reinforcement learning from human feedback (RLHF) \citep{Stiennonetal2020learningto, Baietal2022traininga, Azaretal2024ageneral, Fengetal2024towardsanalyzing} addresses this by using human preference rankings to train a reward model \citep{Christianoetal2017deepreinforcement, Ziegleretal2020fine-tuninglanguage}, followed by reinforcement learning—often with proximal policy optimization (PPO) \citep{Schulmanetal2017proximalpolicy}—to optimize alignment. While effective, RLHF's two-stage process is resource-intensive and sensitive to hyperparameter tuning \citep{Zhengetal2023secretsof}. To overcome these limitations, direct preference optimization (DPO) \citep{Rafailovetal2023directpreference} offers a more streamlined approach by optimizing directly on paired preference data using a classification loss function, eliminating the need for explicit reward model training. Another efficient alternative, group relative policy optimization (GRPO) \citep{Shaoetal2024deepseekmathpushing}—a PPO variant, reduces resource requirements by using group-based scoring to estimate policy advantage. This study provides the first empirical evaluation of DPO and GRPO in the statistics domain. High-quality preference data are crucial for both methods. We employed a powerful external LLM (Gemini) as a teacher model to generate response pairs with clear preference distinctions—an approach inspired by knowledge distillation \citep{Hintonetal2015distillingthe, Calderonetal2023asystematic}. Leveraging a stronger model's judgments provided precise training signals, contributing significantly to our preference optimization success.

Building on these insights, we propose and evaluate a multi-stage training paradigm for developing a domain-specialized statistical LLM. The process begins with the LLaMA-3.2-3B model and incorporates either CoP combined with instruction tuning or SFT using the parameter-efficient LoRA method. These stages are designed to inject core statistical knowledge and task structures into the model. Next, we apply preference alignment using DPO or GRPO to fine-tune the model's outputs according to human preferences. Finally, we perform carefully calibrated downstream task fine-tuning (DTFT) using high-quality question-answering datasets, aiming to improve domain-specific performance while preserving the model's core general capabilities. To assess the effectiveness of this approach, we conducted extensive experiments across several benchmark tasks: GSM8K \citep{Cobbe2021trainingverifiers} for mathematical reasoning, AP Statistics multiple-choice questions \citep{APStat2024apstatistics} for statistical knowledge, and ARC \citep{Clarketal2018thinkyou} for commonsense reasoning.

Key contributions of this study include:
\begin{itemize}
\item Demonstrating the effectiveness of instruction-tuned models as a starting point for domain adaptation;
\item Providing a detailed analysis of performance trade-offs in the SFT stage;
\item Conducting the first comparative study of DPO and GRPO for preference alignment in statistics, highlighting DPO's superior quality and consistency;
\item Quantifying the importance of carefully controlling fine-tuning intensity during DTFT;
\item Delivering a scalable and balanced training paradigm for building high-performance statistical LLMs.
\end{itemize}

The remainder of this paper is organized as follows: Section 2 reviews related work, providing a detailed overview of LLM fine-tuning methods, domain adaptation strategies, instruction-following techniques, alignment mechanisms, and approaches to statistical reasoning. Section 3 introduces the datasets used in this study and how they were leveraged across multiple training and evaluation stages. Section 4 describes our experimental methodology, including implementation details of data preprocessing and the design of the three training pipelines. Section 5 presents experimental results alongside an in-depth comparative analysis of different pipelines and technology combinations. Finally, Section 6 summarizes the key findings, discusses limitations, and outlines promising directions for future research.

\section{Related work}

\subsection{Foundation models}

Foundation models have become a cornerstone of modern NLP, underpinning many state-of-the-art LLMs. These models—such as BERT \citep{Devlinetal2019bertpre-training}, GPT \citep{Radfordetal2018improvinglanguage, Radfordetal2019languagemodels, Brownetal2020languagemodels}, and LLaMA \citep{Touvronetal2023llamaopen, Touvronetal2023llama2open, Grattafiorietal2024thellama3}—are built on the Transformer architecture \citep{Vaswanietal2017attentionis} and are pretrained on massive text corpora using self-supervised learning (SSL) \citep{Jingetal2021self-supervisedvisual}. This pretraining stage enables models to learn deep semantic patterns from unlabeled data, developing broad language understanding and general-purpose reasoning capabilities.

Due to their strong transferability, foundation models can be efficiently adapted to a variety of downstream tasks with minimal additional tuning. However, their general-purpose design often limits performance in specialized domains like statistics, where tasks require precise terminology, domain expertise, and complex reasoning. These limitations highlight the need for targeted adaptation techniques that preserve the model's general reasoning abilities while enhancing its performance in domain-specific contexts.

This study builds on these insights by using the pretrained LLaMA-3.2-3B model as the foundation for a specialized statistical LLM. The model is built on an optimized Transformer decoder-only architecture, featuring several key improvements over the original Transformer design \citep{Ibe2024unlockinglow-resource}. Whereas the traditional Transformer employs a dual-branch encoder–decoder structure with post-layer normalization, LLaMA retains only the decoder and applies pre-normalization using RMSNorm \citep{Zhangetal2019rootmean} at the input of each sub-layer, enhancing stability in deep model training. For positional encoding, LLaMA replaces traditional additive embeddings with rotary position embeddings \citep{suetal2024roformerenhanced}. The self-attention mechanism in LLaMA employs masked attention and supports both multi-head attention and grouped-query attention \citep{Ainslieetal2023gqatraining}, depending on the model scale. LLaMA's feed-forward network adopts the SwiGLU \citep{Shazeer2020gluvariants} activation function. Collectively, these architectural refinements underpin the high efficiency and performance characteristic of the LLaMA family.

\subsection{Supervised fine-tuning}

SFT adapts a pretrained language model to a specific downstream task by learning from labeled data $\mathcal{D}=\{(x,y)\}$, where $x$ is the input and $y$ is the target output. The model parameters $\theta$ are optimized by minimizing the negative log-likelihood (NLL) loss:
\begin{eqnarray} \label{eq:SFTLoss}
 L_{\text{SFT}}(\theta)=-\mathbb{E}_{(x,y) \sim \mathcal{D}}\left[ \sum_{t=1}^{|y|} \log\pi_{\theta}(y_{t}|x, y_{<t}) \right],
\end{eqnarray}
where $\pi_{\theta}(y_{t}|x, y_{<t})$ denotes the probability assigned by the model to token $y_{t}$, given the input $x$ and the preceding tokens $y_{<t}$.

While SFT is often implemented as FFT, updating all model parameters, this approach becomes prohibitively expensive for large-scale LLMs and is susceptible to catastrophic forgetting \citep{Kirkpatricketal2017overcomingcatastrophic, Lopez-Pazetal2017gradientepisodic, Huangetal2024mitigatingcatastrophic}. Earlier work ULMFiT \citep{Howardetal2018universallanguage} introduced techniques to overcome FFT's potential risks. ULMFiT's key strategies included: discriminative fine-tuning (applying different learning rates to different layers, preserving low-level general features), gradual unfreezing (unfreezing layers progressively from top to bottom, allowing adaptation of high-level features before modifying lower-level ones), and slanted triangular learning rates (rapidly increasing the learning rate early in training, then gradually decreasing it to balance adaptation speed with knowledge retention). These methods reduced knowledge degradation and inspired more refined fine-tuning strategies in subsequent large-model training.

\subsection{Parameter-efficient fine-tuning}

The core idea of PEFT is to freeze most of the parameters of a pretrained model $\mathbf{\Phi}_{0}$ and introduce only a small set of additional learnable parameters $\mathbf{\Theta}$, typically accounting for less than 1\% of the total. During fine-tuning, only $\mathbf{\Theta}$ is updated, while $\mathbf{\Phi}_{0}$ remains unchanged. This approach substantially reduces computational and storage costs and mitigates catastrophic forgetting, as the pretrained parameters are preserved, maintaining the general knowledge acquired during pretraining.

Among various PEFT techniques, LoRA \citep{Huetal2021loralow-rank} has emerged as one of the most widely used due to its simplicity and effectiveness. LoRA is based on the hypothesis that the weight update $\triangle W$ to a pretrained matrix $W_{0} \in \mathbb{R}^{d \times k}$ can be decomposed into the product of two low-rank matrices:
\begin{eqnarray*} \label{eq:LoRA1}
 \triangle W=BA,
\end{eqnarray*}
where $B \in \mathbb{R}^{d \times r}$ and $A \in \mathbb{R}^{r \times k}$, with $\text{rank } r \ll \min(d,k)$. The original weight $W_{0}$ remains frozen, $A$ is randomly initialized (e.g., Gaussian), and $B$ is initialized to zero, ensuring that $\triangle W$ is initially zero and does not alter the model's output at the start of training. A scaling factor $\alpha$ controls the adaptation strength, and the updated weights are computed as:
\begin{eqnarray*} \label{eq:LoRA2}
 W^{'}=W_{0}+(\alpha/r)BA.
\end{eqnarray*}
This low-rank approach greatly reduces the number of trainable parameters and achieves performance comparable to, or exceeding, full fine-tuning on many tasks. Furthermore, because $W^{'}$ is merged with $W_{0}$ at inference time, the computational structure remains unchanged, resulting in zero additional inference latency—a key advantage for deployment in latency-sensitive environments.

The success of LoRA has led to numerous extensions. QLoRA \citep{Dettmersetal2023qloraefficient} combines LoRA with 4-bit quantization to reduce memory requirements, enabling fine-tuning of large models on consumer-grade GPUs. DoRA \citep{Liuetal2024doraweight-decomposed} decomposes weights into magnitude and direction, applying low-rank adaptation only to the directional component to potentially improve performance. Further research has examined the impact of LoRA on different layers and modules, such as attention mechanisms and feed-forward networks \citep{Lialinetal2024scalingdown}.

Overall, LoRA and its derivatives demonstrate that PEFT is an indispensable approach in the LLM era. However, while PEFT excels in scenarios with limited task data, it may be less effective when substantial restructuring of a model's domain knowledge is required. In such cases, continual pretraining offers a more comprehensive strategy for injecting specialized knowledge.

\subsection{Continual pretraining}

Supervised fine-tuning alone often cannot provide a model with the depth of domain knowledge required to fully capture specialized terminology, linguistic patterns, and background context. To address this limitation, CoP has been proposed as an effective strategy for domain knowledge enhancement. Its central idea is to continue training a general-purpose pretrained model on large amounts of domain-specific text—typically unlabeled—using SSL objectives. By extending the objectives proven effective in initial pretraining, CoP allows the model to efficiently absorb knowledge from the new corpus.

Two common approaches dominate CoP. Masked language modeling (MLM), popularized by BERT, randomly masks tokens in the input and trains the model to predict them based on context. This promotes bidirectional contextual understanding, crucial for mastering precise domain-specific terminology. Causal language modeling (CLM)/next token prediction (NTP), used in autoregressive models such as GPT, predicts the next token given the preceding sequence, enabling the model to learn domain-specific text generation, terminology sequences, and stylistic coherence. Other SSL variants, such as T5's denoising objective \citep{Raffeletal2020exploringthe} and BART's text infilling \citep{Lewisetal2019bartdenoising}, also provide inspiration for CoP design, though MLM and CLM remain the most widely adopted due to their simplicity and effectiveness.

Recent work has focused on improving CoP's efficiency and stability. For example, the (re)warming strategy leverages better initialization and learning rate schedules to accelerate adaptation and convergence \citep{Guptaetal2023continualpre-training}. The lifelong pretraining framework enables continuous learning from evolving domain corpus while mitigating knowledge forgetting, addressing the challenge of rapidly changing knowledge bases \citep{Jinetal2022lifelongpretraining}.

A key advantage of CoP is that it eliminates the need for costly manual annotation. By immersing the model in large volumes of authentic domain-specific text, CoP enables it to acquire vocabulary, grammar, semantic structures, and even implicit reasoning patterns characteristic of the domain. Empirical evidence shows CoP to be particularly effective in terminology-dense fields, in domains with distinct linguistic styles, or when the target corpus diverges substantially from the original pretraining data \citep{Gururanganetal2020dontstop}.

Despite these strengths, CoP remains fundamentally a language modeling task. It improves a model's ability to predict sequences and absorb knowledge but does not directly provide task execution, instruction following, or preference alignment. For application scenarios requiring explicit task orientation, CoP must therefore be combined with methods such as SFT or instruction tuning. In this sense, CoP builds the ``knowledge foundation'', while fine-tuning strategies determine how effectively the model can apply that knowledge to real-world tasks.

\subsection{Instruction tuning}

Instruction tuning is a pivotal technique that transforms LLMs from simple ``text-completion engines'' into versatile AI assistants capable of understanding user intent and following natural language instructions. Its core mechanism is SFT on large-scale datasets of instruction-input-output pairs that span diverse task types and formats. Through this process, models learn not only to interpret instructions but also to generalize effectively to new, unseen tasks.

Although instruction tuning shares the same training objective as standard SFT (i.e., minimizing the NLL loss (Eq. \ref{eq:SFTLoss})), the difference lies in the data. Training samples are typically formatted as (instruction, input)$\rightarrow$output or (instruction)$\rightarrow$output. Exposure to such data enables models to generate contextually appropriate outputs guided by human intent, significantly enhancing their zero-shot and few-shot learning abilities \citep{Weietal2022finetunedlanguage, Ouyangetal2022traininglanguage, Chungetal2024scalinginstruction-finetuned}. This generalization capacity is central to building flexible and interactive AI systems.

A major factor behind instruction tuning's success is the scale, diversity, and quality of training data. Early approaches reformatted existing NLP datasets—such as question answering, summarization, and translation—into instruction-based formats, unifying hundreds of task types under a single paradigm \citep{Longpreetal2023theflan}. Later, automated data generation methods further advanced scalability. Self-Instruct \citep{Wangetal2023self-Instructaligning} leveraged stronger teacher models to generate new task instructions and responses from a small set of seed instructions, while XL-Instruct \citep{Iyeretal2025xl-instructsynthetic} extended this approach to build large, high-quality corpora through self-guidance. These methods reduce reliance on manual annotation but introduce challenges such as redundancy, lack of diversity, and stylistic uniformity, which necessitate additional control mechanisms. Other research has incorporated real user-model conversation logs into instruction tuning datasets \citep{Taorietal2023alpacaa}. Although limited in scale and sensitive in terms of privacy and compliance, this type of data reflects authentic usage scenarios and enhances a model's responsiveness in open-ended dialogue. Its low cost and high application relevance demonstrate the feasibility of instruction tuning under resource-constrained conditions.

In summary, instruction tuning represents a turning point in LLM development, enabling practical human–computer interaction. Its effectiveness derives not only from fine-tuning strategies but also from the diversity and quality of instruction datasets, which may include reformatted NLP tasks, manual annotations, synthetic data generated by teacher models, and real-world interaction data. Together, these sources equip instruction-tuned models with stronger task comprehension and contextual adaptability, laying a robust ``base policy'' for subsequent alignment techniques such as RLHF.

\subsection{Reinforcement learning from human feedback}

While SFT and instruction tuning enable LLMs to perform tasks and follow natural language instructions, they are insufficient for ensuring alignment with complex human values, ethical standards, and preferences. To address this gap, RLHF has become a widely adopted and highly successful alignment technique. Instead of manually designing reward functions, RLHF derives reward signals indirectly from human preference comparisons over model-generated outputs.

The pipeline of RLHF consists of three stages:
\begin{enumerate}
\item Supervised fine-tuning: A pretrained language model is first fine-tuned on high-quality human demonstrations, producing an initial policy model $\pi_{\text{ref}}$.
\item Reward model training: Human annotators rank or select preferred responses among multiple model outputs $\{y_{1}, y_{2}, \cdots, y_{n}\}$ for a given prompt $x$. These rankings are converted into pairwise comparisons $(x, y_{w}, y_{l})$, where $y_{w}$ is preferred over $y_{l}$. A reward model $\text{RM}_{\phi}$ is trained to assign higher scores to preferred responses using a Bradley–Terry–based objective \citep{Taorietal2023alpacaa}:
    \begin{eqnarray*} \label{BTloss}
     \text{loss}(\phi)=-\mathbb{E}_{(x, y_{w}, y_{l}) \sim \mathcal{D}}\left[ \log\sigma\left( \text{RM}_{\phi}(x, y_{w})-\text{RM}_{\phi}(x, y_{l}) \right) \right].
    \end{eqnarray*}
    Here, $\sigma$ is the sigmoid function, and $\mathcal{D}$ is the preference dataset. Through this process, $\text{RM}_{\phi}$ learns to approximate human judgment and provides a reward signal for policy optimization.
\item Policy optimization: The reward model is then used as feedback for reinforcement learning, where the LLM is treated as a policy $\pi_{\theta}$. In text generation, the state corresponds to the current textual sequence $x_{<i}$, and the action corresponds to the next token $x_{i}$, yielding the policy $\pi_{\theta}(x_{i}|x_{<i})$. The objective is to optimize $\pi_{\theta}$ such that generated sequences maximize cumulative rewards from $\text{RM}_{\phi}$.

    A challenge arises when optimizing solely for $\text{RM}_{\phi}$ rewards: the model may diverge excessively from the reference distribution $\pi_{\text{ref}}$, a phenomenon known as reward hacking. To counter this, RLHF incorporates a KL-divergence penalty that constrains the new policy to remain close to $\pi_{\text{ref}}$. The adjusted reward signal becomes:
    \begin{eqnarray*} \label{adjustedRM}
     R^{'}(x_{<i+1})=R(x_{<i+1})-\beta D_{\text{KL}}\left( \pi_{\theta}(\cdot|x_{<i}) \ \| \ \pi_{\text{ref}}(\cdot|x_{<i}) \right)
    \end{eqnarray*}
    where $R(x_{<i+1})$ is the sequence-level reward from $\text{RM}_{\phi}$, $\beta$ controls the KL penalty strength, and $D_{\text{KL}}$ measures divergence between the two policies. This formulation ensures that policy optimization balances alignment with human preferences and preservation of natural language distribution.
\end{enumerate}

PPO \citep{Schulmanetal2017proximalpolicy, Zhengetal2023secretsof} is the most widely used reinforcement learning algorithm in RLHF, valued for its stability and efficiency. It employs importance sampling to reuse data generated by an old policy and introduces a clipping mechanism that constrains policy updates, preventing large deviations and stabilizing training. This balance of sample efficiency and stability makes PPO particularly well-suited for large language model alignment. Despite its success in aligning LLMs \citep{Devlinetal2019bertpre-training, Radfordetal2018improvinglanguage, Touvronetal2023llamaopen}, PPO remains complex, resource-intensive, and sensitive to hyperparameters. Challenges include designing robust reward models, preventing the pattern collapse (instability in policy learning) \citep{Zhengetal2023secretsof}, and mitigating the alignment tax (reduced performance on traditional NLP tasks) \citep{Linetal2024mitigatingthe}.

To address these issues, alternative preference optimization methods have been proposed. DPO \citep{Rafailovetal2023directpreference} simplifies training by bypassing the reward model and learning directly from preference pairs with a classification-style objective. It derives its loss function from the KL-regularized RLHF objective and uses a closed-form reparameterization based on the Bradley-Terry model. Advantages of DPO include: no need for a reward model or value network, greater training stability, and lower computational cost. However, DPO may be sensitive to the quality of the SFT model and can exhibit asymmetric learning dynamics when preference margins are small \citep{Pan2025etalwhatmatters}.

GRPO \citep{Shaoetal2024deepseekmathpushing}, a lightweight PPO variant, eliminates the value network and estimates the advantage function using group-average reward baselines. It retains PPO's clipping mechanism and KL regularization, providing stable policy updates with reduced memory consumption. GRPO is particularly suited for applications with limited compute resources, high-throughput preference training, and step-by-step supervision needs.

\subsection{Knowledge distillation}

Knowledge distillation (KD) is a model compression technique designed to transfer knowledge from a large, high-performing teacher model (T) to a smaller, more efficient student model (S) \citep{Hintonetal2015distillingthe}. The objective is to reduce inference costs without significantly sacrificing performance, making models more suitable for deployment on resource-limited or latency-sensitive systems.

In natural language generation, KD can be applied at different levels. For word-level KD (logits KD), the student is trained to match the teacher's probability distribution over the next token at each time step by minimizing the KL divergence between them \citep{Calderonetal2023asystematic}. This allows the student to capture fine-grained linguistic and semantic knowledge but requires direct access to the teacher's logits—often infeasible when the teacher is a closed-source API. For sequence-level KD, instead of imitating token-level distributions, the student learns from full output sequences $\hat{y}_{T}$ generated by the teacher for inputs $x$. These sequences act as pseudo-targets \citep{Calderonetal2023asystematic, Shleiferetal2020pre-trainedsummarization}. This approach is more practical in black-box settings since it requires only the teacher's final outputs, not intermediate logits.

In this study, we adopt sequence-level KD. Using the teacher model's API, we generated a large set of pseudo-labeled pairs $(x, \hat{y}_{T})$, which serve as training data for SFT on the student model. The objective minimizes the NLL Loss (Eq. \ref{eq:SFTLoss}) of the student when predicting pseudo-targets. The main advantage of this method is its simplicity and practicality: it requires only teacher outputs, not internal states or logits. Furthermore, pseudo-labeled data from KD has proven valuable in reinforcement learning alignment pipelines, providing supervision signals for reward model training or policy optimization. This enables the construction of competitive preference models without costly human annotation, making KD a powerful complement to RLHF.

\subsection{Downstream task fine-tuning}

Even after RLHF alignment, model preferences often remain biased toward general-purpose language generation. To optimize performance for specific applications, a final stage of DTFT is applied. This process uses a relatively small amount of task-specific labeled data to make fine adjustments to the aligned model, ensuring that outputs better satisfy domain requirements and evaluation criteria.

Recent studies \citep{Hsiehetal2023distillingstep-by-step} highlight that DTFT not only improves task adherence but can also transfer more complex capabilities, thereby enhancing the adaptability of smaller models in specialized, task-oriented applications. In practice, sequence-level knowledge distillation often serves as an effective initialization or data augmentation strategy to quickly build a student model with basic language understanding and generation capabilities, but its full potential is realized only when combined with DTFT.

\section{Materials}

\subsection{Dataset collection}
\label{ss:used-dataset}

This study used a variety of datasets for developing and evaluating a large language model specialized in statistics. A complete overview of dataset names, sources, descriptions, and sizes is provided in Table \ref{tab:dataset}. To build a strong foundation in statistics, we collected large-scale corpora, primarily for CoP, aiming to inject background knowledge, terminology, and linguistic patterns into the model. For instruction tuning, SFT, and DTFT, we collected and generated diverse instruction/question–answer (QA) datasets. These datasets trained the model to understand and follow instructions, perform QA, and execute specific tasks. To ensure alignment with human preferences, we curated pairwise preference datasets, with portions of both preference and QA data generated through carefully crafted prompts to the Gemini API. Finally, to evaluate performance, we employed a suite of benchmarks covering multiple dimensions. Together, this diverse evaluation suite allows us to analyze both the trade-offs between general reasoning and professional expertise, and the overall effectiveness of the domain adaptation process.

\begin{sidewaystable}[htbp]
\footnotesize
\centering
\caption{Overview of datasets collected in this study}
\label{tab:dataset}
\renewcommand{\arraystretch}{1.2}
\begin{tabular}{p{4.5cm} p{4.5cm} p{8.0cm} p{2.0cm}}
\hline
\textbf{Dataset name} & \textbf{Source} & \textbf{Description} & \textbf{Size} \\
\hline

\multicolumn{4}{l}{\textbf{Large corpora}} \\
\texttt{arXiv (stat.*)} & arXiv & Full-text papers in statistics & 29,703 \\
\texttt{S2ORC} & S2ORC & Academic paper paragraphs filtered by statistical keywords & 7,496 \\
\texttt{Statistics Open-Source Books} & Springer Nature DOAB & Open-access statistics books & 105 \\
\texttt{Statistics Course Books} & OpenStax & University-level statistics textbooks & 21 \\
\hline

\multicolumn{4}{l}{\textbf{Instruction/Question–Answer (QA)}} \\
\texttt{OpenHermes 2.5} & Hugging Face & General-purpose instruction-following dataset & 100,000 \\
\texttt{Dolly-15k} & Hugging Face & General-purpose instruction-following dataset & 15,000 \\
\texttt{Statistical Nouns/Defs} & Wiki & Statistical terms with definitions & 1,203 \\
\texttt{Statistical CoT} & Gemini API & Chain-of-thought (CoT) QA pairs in statistics & 1,207 \\
\texttt{FineTome-100k} & Hugging Face & High-quality supervised fine-tuning dataset & 20,000 \\
\texttt{Math-QA} & Kaggle & Math problem QA dataset & 10,000 \\
\texttt{Cross Validated} & StackExchange & Statistics QA pairs (community-validated, votes>10) & 9,071 \\
\texttt{Knowledge Graph} & Wiki/STATO/OBCS & QA pairs constructed from knowledge graph concepts in statistics & 8,414 \\
\texttt{Statistical Conversation} & Gemini API & Multi-turn conversations on statistical reasoning and concepts & 1,054 \\
\texttt{GSM8K (Train)} & OpenAI/Hugging Face & Grade-school math word problems (training set only, for model training) & 7,473 \\
\hline

\multicolumn{4}{l}{\textbf{Preference data}} \\
\texttt{Statistical GRPO} & Gemini API & Preference pairs for GRPO training & 2,255 \\
\texttt{Statistical DPO} & Gemini API & Preference pairs for DPO training & 2,382 \\
\texttt{Math DPO} & Hugging Face & Preference pairs for DPO training & 2,393 \\
\hline

\multicolumn{4}{l}{\textbf{Benchmark}} \\
\texttt{GSM8K (Test)} & OpenAI/Hugging Face & Grade-school math problems (test set reserved for 8-shot CoT evaluation) & 1,319 \\
\texttt{ARC} & Allen Institute for AI & Commonsense reasoning benchmark (0-shot evaluation on Easy Set) & 2,357 \\
\texttt{AP Statistics} & AP Central & College-level introductory statistics multiple-choice questions & 287 \\
\hline

\end{tabular}
\normalsize
\end{sidewaystable}

Not all collected datasets were used in the core training pipeline. Some, such as \texttt{arXiv (stat.*)}, and statistics open-source books, were excluded after evaluation due to quality concerns, limited relevance, or resource constraints. Nonetheless, collecting and analyzing these datasets provided valuable insights for data selection strategy in this study.

\subsection{Construction of training datasets}
\label{ss:training-datasets}

To effectively leverage diverse datasets across multiple training stages, this study adopted a systematic dataset construction and integration framework. This process involved not only the cleaning and normalization of raw data (as discussed in Section \ref{ss:data-preprocessing}) but also the design of data organization methods, mixing strategies, prompt engineering, and format conversions tailored to specific training objectives (CoP, instruction tuning, SFT, RLHF, and DTFT). These strategies aimed to maximize data utility, ensure effective skill acquisition at each stage, and optimize model learning efficiency under constrained computational resources.

\subsubsection{Continual pretraining}
\label{sss:training-datasets_CoP}

The goal of the CoP stage was to immerse the model in a professional statistical language environment, enabling it to internalize domain-specific terminology, writing conventions, and conceptual knowledge. Two primary datasets were used: academic paper excerpts (\texttt{S2ORC}) and statistical terms and definitions (\texttt{Statistical Nouns/Defs}). The \texttt{S2ORC} subset was precisely filtered using title and subtitle keywords to ensure their relevance to statistical domain. Each selected paper was stored as a dictionary, with the paper title and keywords as the key and paragraph excerpts as values. Structured fields such as ``Problem'', ``Method'', ``Result'', and ``Conclusion'' were then extracted to form the final dataset. To strengthen understanding of key terminology, the \texttt{Statistical Nouns/Defs} data were repeated five times, whereas academic paper excerpts were used once in the training set. The combined dataset was trained for two full epochs.

Apart from the direct use of text corpora in the CoP stage, in the subsequent stages of instruction tuning, SFT, RLHF, and DTFT, all data were format-converted through a default chat template to train the model's understanding and generation abilities in conversational interaction scenarios.

\subsubsection{Instruction tuning}

The instruction tuning stage established the model's foundational instruction-following capability. Two large-scale general-purpose instruction-following datasets: \texttt{Dolly-15k} and \texttt{OpenHermes 2.5} were used, each converted into the chat format \texttt{[\{"role": "user", "content": "..."\}, \{"role": "assistant", "content": "..."\}]}. Neither dataset was augmented or repeated; the combined corpus was trained for one epoch. This setup provided broad exposure to general instruction patterns while avoiding overfitting or premature bias toward specific response formats. The primary objective was to quickly build a general instruction-response framework rather than to perform deep optimization on any specific general skill.

\subsubsection{Supervised fine-tuning}

The SFT stage enhanced the model's reasoning, statistical problem-solving, and complex instruction-handling abilities through a carefully constructed mixture of datasets with differentiated augmentation:
\begin{itemize}
\item \texttt{GSM8K (Train)}: step-by-step mathematical reasoning (no augmentation)
\item \texttt{Statistical CoT}: chain-of-thought reasoning, repeated 2$\times$
\item \texttt{Statistical Nouns/Defs}: term definitions, repeated 3$\times$
\item \texttt{Statistical GRPO}: preferred response style examples used here as supervised signals, repeated 2$\times$
\end{itemize}
All subsets were merged into a single mixed dataset and trained for three epochs. This differentiated augmentation ensured that more complex or domain-intensive tasks were represented more frequently, allowing the model to develop deeper reasoning and contextual understanding while maintaining balance through \texttt{GSM8K}'s unaltered baseline proportion.

In addition, two further datasets: \texttt{Math-QA} (mathematical problem QA) and \texttt{FineTome-100k} (general-purpose supervised fine-tuning data) were incorporated in variant configurations to examine how different SFT setups influence overall performance.

\subsubsection{Reinforcement learning from human feedback}

The RLHF stage aimed to align the model's output with human preference signals through contrastive learning on pairwise preference data. Three datasets were used: \texttt{Statistical GRPO}, \texttt{Statistical DPO}, and \texttt{Math DPO}. In training DPO version of RLHF, \texttt{Statistical DPO} and \texttt{Math DPO} datasets were unaugmented and combined for two training epochs. By jointly training on preference data from both the mathematics and statistics domains, the model learned preference signals across different task types. This approach enabled it to internalize not only general response quality patterns but also the domain-specific stylistic preferences unique to each field, thereby enhancing its ability to produce precise, human-aligned responses in both mathematical and statistical tasks.

\subsubsection{Downstream task fine-tuning}

DTFT served as the final refinement stage, adapting the model to high-quality, task-specific data closely aligned with real-world applications. This stage employed:
\begin{itemize}
\item \texttt{Cross Validated}: real QA pairs from professional statistics forums
\item \texttt{Knowledge Graph}: concept-based QA generated from knowledge graphs
\item \texttt{Statistical Conversation}: multi-turn domain dialogues
\end{itemize}
All datasets were combined without augmentation and trained for one epoch. This final fine-tuning focused on style optimization, knowledge calibration, and format adaptation rather than large-scale capability reshaping, thereby improving real-task performance while minimizing the risk of catastrophic forgetting.

\subsubsection{Summary of dataset construction strategy}

In summary, the dataset design and mixing strategies at each training stage were guided by a holistic consideration of data characteristics, training objectives, model learning dynamics, and resource constraints. Repetition factors and epoch counts were systematically adjusted to balance data influence and computational efficiency. This structured approach ensured comprehensive enhancement of the model’s performance across all key dimensions—statistical knowledge acquisition, reasoning capability, instruction following, and alignment with human preferences—while maintaining efficiency and stability throughout the training pipeline.

\subsection{Model evaluation benchmarks}
\label{ss:benchmarks}

To comprehensively evaluate the performance of the statistical LLM developed in this study, an integrated evaluation framework was established. This framework was designed to address the limitations of conventional language model metrics, which often fail to capture the depth of statistical knowledge, reasoning accuracy, and domain-specific understanding. The evaluation system combines quantitative monitoring metrics on standardized benchmark tests and a custom-built domain-specific evaluation dataset, and qualitative expert analysis.

During training, real-time monitoring of cross-entropy loss, learning rate, and gradient norm was performed to track the model's learning progression and stability. For formal performance assessment, the model was evaluated using the following benchmarks and methods.

\subsubsection{Grade School Math 8K (GSM8K)}

GSM8K, released by OpenAI, is a widely used benchmark for assessing multi-step mathematical reasoning. It contains approximately 8,500 natural language math word problems, each paired with a complete step-by-step solution. This study adopted the 8-shot CoT evaluation paradigm (\texttt{GSM8K (Test)}), in which the model receives eight annotated examples of ``question-reasoning-answer'' before being asked a new question. This setup encourages the model to emulate step-by-step logical reasoning during inference. The primary evaluation metric is answer accuracy, reflecting the model's capability in logical reasoning and numerical computation.

\subsubsection{AI2 Reasoning Challenge (ARC)}

Developed by the Allen Institute for AI (AI2), the \texttt{ARC} benchmark evaluates common-sense reasoning using questions derived from U.S. grade 3-9 science examinations. To investigate whether domain-specific training in statistics affects general reasoning ability, this study conducted a zero-shot evaluation on the ARC-Easy Set, where no examples are provided prior to questioning. By analyzing changes in accuracy on this benchmark, we can determine whether specialized training introduces interference or whether the model maintains its general reasoning competence alongside its domain specialization.

\subsubsection{AP Statistics Benchmark}

Given the absence of public benchmarks tailored to testing core statistical competence, we constructed a dedicated \texttt{AP Statistics} benchmark. The dataset was sourced primarily from publicly available question bank CrackAP.com's AP Statistics Practice Tests, initially containing 678 multiple-choice questions. After manual curation to remove items containing images or unsuitable formatting, 287 representative text-only questions were retained. All questions were standardized into a uniform format, each consisting of a question stem and five options labeled (A)-(E), and stored in JSON format for reproducibility.

This benchmark evaluates the model's understanding and application of fundamental statistical principles, including descriptive statistics, probability distributions, sampling and estimation, hypothesis testing, and regression analysis. For assessment, a custom prompt instructed the model to output only the letter corresponding to the correct choice, with regular expressions to automatically extract the answer from the model output. The final evaluation metric was accuracy, providing a reliable, quantitative measure of statistical domain expertise.

\subsubsection{Qualitative evaluation}

Beyond quantitative benchmarks, qualitative expert evaluation was conducted to provide a deeper assessment of the model's conceptual understanding and expressive precision. We selected a set of representative open-ended questions probing core areas of statistical reasoning—such as the interpretation and misconceptions of confidence intervals, the distinction between correlation and causation, and the proper use and misinterpretation of p-values in hypothesis testing. Appendix Table \ref{atab:model_samples} includes some representative examples that show the performance differences between different models.

Responses from models at different training stages (e.g., an instruction-tuned baseline versus a statistics-specialized model) were compared and reviewed by human experts. Evaluation focused on conceptual accuracy, logical coherence, clarity of explanation, appropriate use of terminology, and ability to recognize or avoid statistical fallacies. This expert-reviewed qualitative analysis complements quantitative metrics by offering high-resolution insights into the model's reasoning structure, interpretive accuracy, and linguistic expressiveness.

\section{Methods}

\subsection{Data preprocessing}
\label{ss:data-preprocessing}

Transforming raw, heterogeneous data into high-quality input suitable for LLM training is a foundational step in building a domain-specific model. This process demands not only precise technical implementation but also a careful understanding of diverse data characteristics, potential pitfalls, and effective processing strategies.

In this study, we worked with a wide variety of statistics-related datasets and conducted extensive technical explorations in preprocessing and engineering. Some approaches were ultimately excluded from the core pipeline due to suboptimal empirical results, limited efficiency, or incompatibility with the target model architecture. Nevertheless, these explorations provided valuable insights for refining our preprocessing strategy.

The following subsections outline the specific practices and technical considerations applied in this study to prepare heterogeneous data for training and alignment.

\subsubsection{Basic general processing}

The first stage of all preprocessing workflows involves cleaning and standardizing raw text. The goal is to establish a clean, consistent foundation for downstream processing and training. Using standard regular expressions and string operations, we removed residual web artifacts (e.g., HTML/XML tags) and normalized whitespace, newlines, and tab characters. Text encodings were converted to UTF-8 to avoid character-set inconsistencies.

To address structural noise from PDFs and web-crawled sources (such as repeated headers, footers, and page numbers), we designed a pattern recognition mechanism for automatic filtering. For academic texts, reference sections were removed using keyword detection (e.g., ``References'', ``Bibliography'') to increase corpus knowledge density. Finally, while narrative text was lowercased for consistency, statistical terms and mathematical symbols were preserved in their original form.

\subsubsection{Processing of PDF data}

PDF data constituted a large portion of our corpora, including \texttt{Statistics Open-Source Books}, \texttt{Statistics Course Books}, and \texttt{arXiv (stat.*)}. Due to their unstructured nature, we designed a specialized workflow. Text and block coordinates were extracted using PyMuPDF (fitz) \citep{McKie2024pymupdfa}, forming the basis for semantic reconstruction. We also experimented with combining title keywords and block positions to identify chapter boundaries, particularly in arXiv papers. However, due to significant variability in document layouts, this method could not consistently and accurately segment chapters and was therefore excluded from the main pipeline.

For multimodal processing, we explored using the BLIP model \citep{Lietal2022blipbootstrapping} to generate textual descriptions of statistical charts to supplement image content. Yet, because of the model's limited accuracy in interpreting professional charts and the difficulty of reliably matching figures with annotations, these descriptions were not incorporated into the core workflow. For mathematical formulas, we evaluated Pix2Text \citep{Breezedeus2024pix2textan} for converting image-based equations into LaTeX. While reasonably effective for simple formulas, it struggled with complex layouts and introduced speed and compatibility issues. We therefore retained formulas in their original text style, allowing the model to infer their meaning from surrounding context.

Finally, to mitigate duplication caused by page breaks or formatting artifacts, we implemented a similarity detection mechanism based on Levenshtein distance \citep{Levenshtein1965binarycodes}, which identified and removed repeated content in adjacent paragraphs, thereby improving corpus purity and training efficiency.

\subsubsection{Processing of structured sources}

Beyond plain text, we incorporated structured knowledge sources, such as the STATO \citep{STATO2025} and OBCS \citep{Zhengetal2016theontology} ontologies. These provide explicit semantics and relationships in a knowledge graph (KG) format. Using pandas \citep{reback2020pandas}, we cleaned the raw data and built a mapping dictionary linking ontology entity URIs to their preferred labels. We then iterated through triplets (subject, predicate, object) to generate standardized text statements. To improve relational coverage, inverse triplets were automatically added following predefined rules. Duplicates were removed via set operations, yielding a clean collection of structured statements suitable for model learning and integration into stages such as SFT.

\subsubsection{Extracting structured knowledge from text}

As a supplementary exploration, we attempted to extract structured semantic knowledge directly from unstructured academic text. Using SpaCy's en\_core\_web\_lg model \citep{Honnibaletal2020spacyindustrial-strength} with dependency parsing, we extracted subject-verb-object (SVO) triplets. In principle, this approach could support automatic construction of a semantic network. However, the complexity of statistical writing—long sentences, subordinate clauses, and technical terminology—led to low accuracy and limited coverage compared to curated ontologies. As such, these results were excluded from the main training corpus and retained only for auxiliary analysis.

\subsubsection{Unified formatting and tokenization}

Once cleaning and structural modeling were complete, all text was converted into numerical sequences for training. We adopted a pre-tokenization strategy, ensuring all text was tokenized and ID-mapped prior to training, which reduced overhead and improved efficiency. Tokenization was performed using Hugging Face's AutoTokenizer \citep{AutoTokenizer2020}, configured to match the vocabulary and algorithm of the base model. Texts exceeding the model's maximum sequence length were processed using a sliding window approach to preserve context, while shorter inputs were padded. Padding tokens were assigned a label value of –100, ensuring they did not affect loss computation.

\subsubsection{Prompt engineering}

Prompt engineering played a central role throughout data preparation, training, and evaluation. The design quality of prompts directly influenced model learning efficiency and downstream performance. To address different tasks and training stages, we developed multiple prompt templates, including:
\begin{itemize}
\item Prompts for creating CoT QA pairs in statistics.
\item Prompts for producing preference samples (question, reasoning, answer) for GRPO and (prompt–chosen–rejected) for DPO training.
\item Prompts for generating multi-turn conversations on statistical reasoning and concepts for DTFT.
\item Benchmark-specific evaluation prompts to ensure fairness and comparability.
\end{itemize}
All prompt templates were applied after tokenization, converting the data into structured inputs with standardized response formats. Through iterative refinement, carefully designed prompts substantially improved the model's ability to generate reliable, task-specific outputs, thereby enhancing both performance and applicability in statistics. The full set of prompt templates is provided in Appendix Table \ref{atab:prompt}.

\subsection{Model architecture and training framework}
\label{ss:model-architecture}

\subsubsection{Foundation models and parameter-efficient fine-tuning}

This study adopts Meta's LLaMA-3.2-3B as the base foundation model for all fine-tuning and experimental procedures. The model is built on an optimized Transformer decoder-only architecture, featuring several key improvements over the original Transformer design \citep{Ibe2024unlockinglow-resource}. LLaMA-3.2-3B was selected for this study due to its open availability, competitive performance, and excellent compatibility with PEFT methods.

PEFT in this study primarily employed LoRA, which introduces low-rank trainable matrices into targeted model layers, enabling efficient adaptation with minimal parameter updates. LoRA's rank and scaling factor ($\alpha$) were adjusted dynamically based on task type and training stage. To further reduce memory consumption and computation costs, the base model was loaded using 4-bit quantization via the bitsandbytes library \citep{Dettmersetal2023bitsandbytes}. Training employed the AdamW optimizer, with gradient accumulation used to achieve an effective batch size appropriate for available hardware resources.

\subsubsection{Training and fine-tuning framework}

To accommodate both local and cloud-based resource environments, the overall experimental workflow was built on the Hugging Face ecosystem. For small- to medium-scale GPU setups, we employed the Unsloth acceleration library \citep{Hanetal2023unsloth}, which is optimized for PEFT methods such as LoRA. By leveraging optimized CUDA kernels and efficient memory management, Unsloth reduces VRAM usage by approximately 20–30\% and increases training speed by 1.5–2$\times$ compared with the standard Hugging Face Trainer, based on preliminary testing.

For large-scale training during the CoP stage or for distributed multi-GPU fine-tuning, we integrated the DeepSpeed framework \citep{Rasley2020deepspeedsystem}. Developed by Microsoft, DeepSpeed implements advanced distributed optimization strategies, most notably the ZeRO (zero redundancy optimizer), which shards model parameters, gradients, and optimizer states across multiple GPUs, thereby minimizing memory overhead. This study utilized ZeRO Stage 2 with 8 A100 GPUs provided by Taiwan Computing Cloud, significantly accelerating training throughput. While DeepSpeed offers excellent scalability, integration required careful configuration and compatibility tuning with the Hugging Face Trainer, along with additional engineering considerations such as loss monitoring and runtime stability.

\subsection{Core experimental method: Multi-stage training strategy}
\label{ss:method}

To address the challenges of domain adaptation for LLMs and to establish an efficient framework for developing a high-performance statistical LLM, this study systematically designed and compared three representative multi-stage training pipelines. Each pipeline integrates different sequences and combinations of techniques to examine how training order and stage composition affect model performance and adaptation efficiency.

All pipelines used the constructed training datasets described in Section \ref{ss:training-datasets}, were implemented under the training framework outlined in Section \ref{ss:model-architecture}, and were evaluated using the benchmarks and metrics detailed in Section \ref{ss:benchmarks}.

\subsubsection{Pipeline 1 — Knowledge-first strategy}

\begin{figure}
 \begin{center}
  \includegraphics[width=\textwidth]{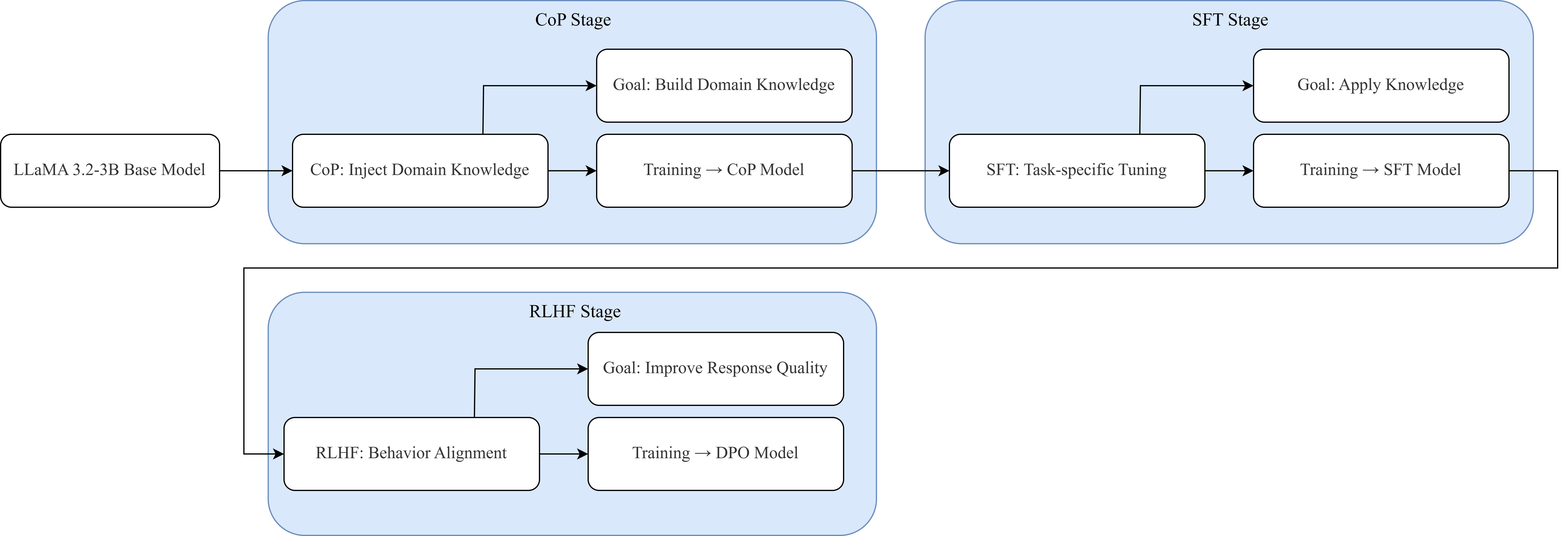}
 \end{center}
\caption{Flowchart of Pipeline 1.}
\label{fig:pipeline1}
\end{figure}

Pipeline 1 (Figure \ref{fig:pipeline1}) prioritizes early domain knowledge injection. The underlying hypothesis is that exposing the model to a large corpus of statistical texts before any instruction-based fine-tuning allows it to deeply internalize domain-specific concepts, terminology, and discourse structure. This immersion is expected to establish a strong foundation in statistical reasoning and linguistic patterns, thereby supporting later task-oriented fine-tuning.

Implementation begins with the LLaMA-3.2-3B Base model, which undergoes CoP using a mixed statistical corpus. This corpus, comprising academic paper paragraphs (\texttt{S2ORC}) and core statistical terminology and definitions (\texttt{Statistical Nouns/Defs}), serves to enhance the model's grasp of professional vocabulary and conceptual relationships. After CoP, the model proceeds to SFT using structured statistical instruction and question–answer data, allowing it to apply the learned knowledge in interactive, task-based contexts. Finally, DPO version of RLHF is applied to align the model's outputs with human judgment, leveraging expert-reviewed statistical preference data. This step enhances the clarity, accuracy, and professionalism of responses. Pipeline 1, therefore, represents a ``knowledge-first'' paradigm, evaluating whether early domain immersion—prior to SFT—can yield stronger specialization and whether it introduces limitations in later task learning performance.

\subsubsection{Pipeline 2 — Instruction-bridge strategy}

\begin{figure}
 \begin{center}
  \includegraphics[width=\textwidth]{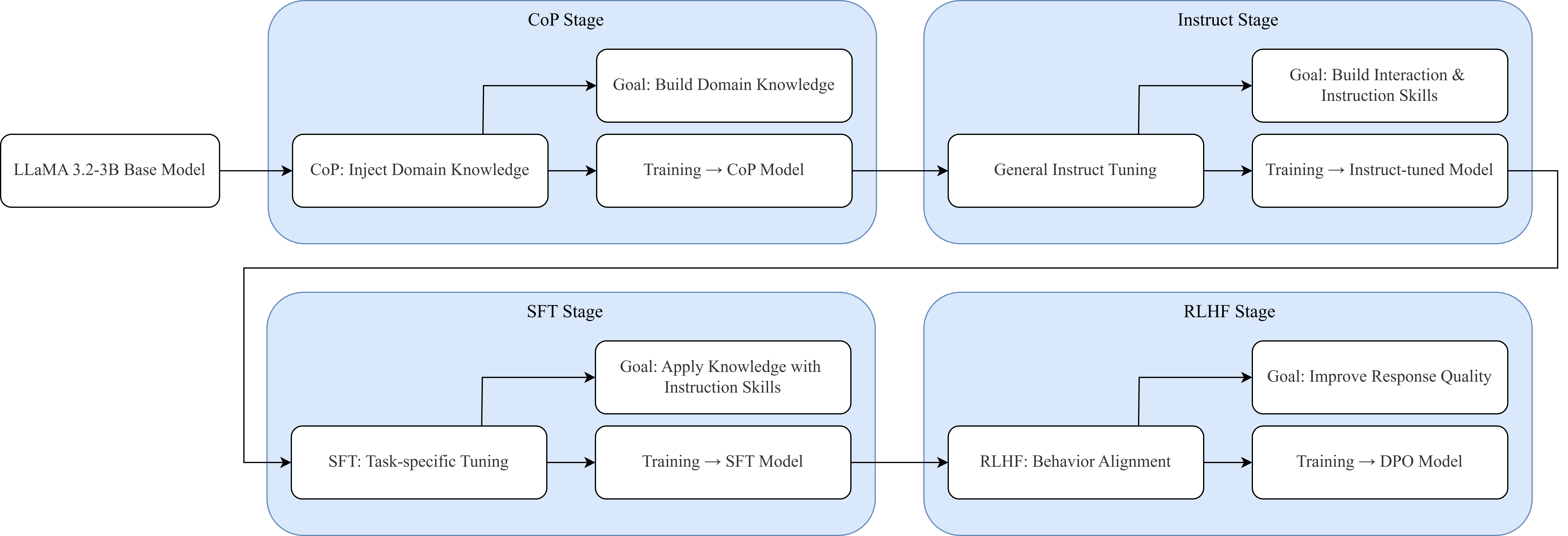}
 \end{center}
\caption{Flowchart of Pipeline 2.}
\label{fig:pipeline2}
\end{figure}

Pipeline 2 (Figure \ref{fig:pipeline2}) was designed to address a key limitation of Pipeline 1—its potential deficiency in instruction-following ability. While early domain training builds knowledge depth, the model still requires basic interaction and comprehension skills to effectively apply that knowledge. Hence, Pipeline 2 introduces an instruction tuning stage as an intermediary between CoP and SFT.

Implementation begins identically to Pipeline 1, with CoP performed on the same statistical corpus to establish foundational domain knowledge. Next, an instruction tuning phase using open-source, general-purpose instruction datasets (e.g., \texttt{Dolly-15k} and \texttt{OpenHermes-2.5}) equips the model with broad interaction and command-following capabilities. Subsequently, the model undergoes domain-specific SFT, integrating statistical reasoning, question answering, and task-based learning. The final RLHF-DPO stage further refines the model's responses, improving fluency and adherence to statistical reasoning norms.

This pipeline investigates the synergistic relationship between early knowledge acquisition and general instruction capability. Specifically, Pipeline 2 tests whether interleaving general instruction tuning before domain-specific fine-tuning leads to superior adaptation and efficiency compared to Pipeline 1. Additionally, it provides a basis for comparison against models that begin with pre-existing instruction-following abilities.

\subsubsection{Pipeline 3 — Instruction-first domain adaptation}

\begin{figure}
 \begin{center}
  \includegraphics[width=\textwidth]{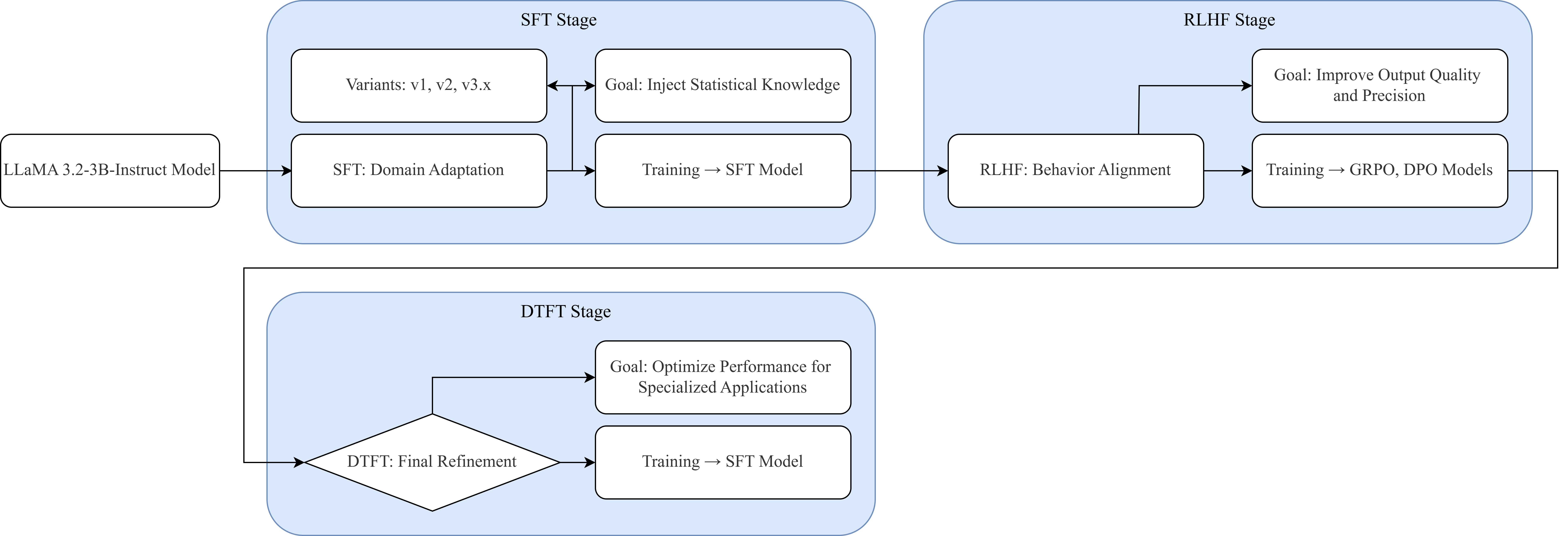}
 \end{center}
\caption{Flowchart of Pipeline 3.}
\label{fig:pipeline3}
\end{figure}

Pipeline 3 (Figure \ref{fig:pipeline3}) represents the core experimental direction of this study and diverges fundamentally from the previous two strategies. Instead of starting from a base model, Pipeline 3 builds upon LLaMA-3.2-3B-Instruct, a model that has already undergone large-scale instruction fine-tuning and possesses robust general reasoning and instruction-following abilities. The hypothesis underlying this strategy is that a model with strong general-purpose capabilities may require only targeted domain adaptation to achieve expert-level performance. This could yield comparable or superior results with reduced resource consumption and training time.

Training begins with LLaMA-3.2-3B-Instruct's SFT on statistical tasks to orient the model toward the domain's specific concepts, data structures, and reasoning patterns. Next, RLHF is used for preference alignment to enhance response quality and domain accuracy. To further enhance the model's adaptability to real-world applications, an DTFT stage can be applied, allowing for precision adjustment depending on experimental goals. Within the SFT stage, multiple sub-versions were implemented to examine how different configurations—such as data composition, fine-tuning length, and parameter-efficient tuning strategies (e.g., training only response tokens)—affect performance. Similarly, RLHF preference optimization methods were compared between GRPO and DPO.

Pipeline 3 thus aims to validate the feasibility and efficiency of domain adaptation based on a high-performance instruction model. Through comparative analysis with Pipelines 1 and 2, it evaluates trade-offs in training cost, domain knowledge retention, and task effectiveness, providing insights into optimal strategies for building specialized LLMs under resource constraints.

\section{Experimental results}

This section presents and analyzes the results of the three multi-stage training pipelines introduced in Section \ref{ss:method}. The overarching goal is to evaluate how different combinations of knowledge injection, instruction tuning, supervised fine-tuning, and preference optimization contribute to constructing a high-performance statistical LLM. A key focus is determining how to effectively integrate statistical-domain knowledge into the LLaMA-3.2-3B architecture while maintaining—or even enhancing—its general reasoning capabilities.

To establish a clear baseline, we begin by reporting the performance of the LLaMA-3.2-3B Base model and its Instruct variant on the three core evaluation benchmarks: \texttt{GSM8K (Test)} (8-shot + CoT), \texttt{AP Statistics} (0-shot), and \texttt{ARC} (0-shot).
\begin{itemize}
\item Base—accuracy \%: 29.87 (\texttt{GSM8K (Test)}), 26.13 (\texttt{AP Statistics}), 42.24 (\texttt{ARC})
\item Instruct—accuracy \%: 64.44 (\texttt{GSM8K (Test)}), 37.63 (\texttt{AP Statistics}), 43.60 (\texttt{ARC})
\end{itemize}
As expected, instruction tuning yields significant gains in mathematical reasoning, general reasoning, and instruction following. These baseline results serve as essential reference points for interpreting the effectiveness of the three training pipelines.

\subsection{Pipeline 1: Limitations of direct domain knowledge injection}

Pipeline 1 begins with the LLaMA-3.2-3B Base model and prioritizes domain-specific CoP, operating under the hypothesis that early exposure to large amounts of statistics text would establish a strong knowledge foundation. The benchmark results are shown in Table \ref{tab:pipeline1_results}.

\begin{table}[htbp]
\footnotesize
\centering
\caption{Benchmark results for training Pipeline 1 (knowledge-first strategy).}
\label{tab:pipeline1_results}
\renewcommand{\arraystretch}{1.2}
\begin{threeparttable}
\begin{tabular}{lccc}
\toprule
\textbf{Model} & \textbf{GSM8K (Test)} & \textbf{AP Statistics} & \textbf{ARC} \\
 & (8-shot + CoT, Acc.\ \%) & (0-shot, Acc.\ \%) & (0-shot, Acc.\ \%) \\
\midrule
LLaMA-3.2-3B                & 29.87          & 26.13          & 42.24 \\
\hline
+ CoP                       & 29.49          & 26.13          & 42.24 \\
+ CoP + SFT                 & \textbf{33.89} & 26.13          & \textbf{43.52} \\
+ CoP + SFT + RLHF          & 27.90          & \textbf{26.48} & 43.34 \\
\bottomrule
\end{tabular}
\begin{tablenotes}
\item Values in boldface denote the best result for each benchmark.
\end{tablenotes}
\end{threeparttable}
\normalsize
\end{table}

The results indicate that Pipeline 1 is largely ineffective. Despite extensive exposure to statistics text, CoP produced no meaningful improvement on any benchmark. All scores remained near baseline. This suggests that passive knowledge exposure—without task-guided objectives—does not lead to improved statistical reasoning or conceptual understanding. The model gains shallow familiarity with the domain's vocabulary but does not develop usable competencies.

SFT yields a modest improvement on \texttt{GSM8K (Test)} (+4\%), likely reflecting the benefit of structured QA data. However, the \texttt{AP Statistics} score remains unchanged, indicating that the model still cannot apply the statistical text it observed during CoP to domain-specific problem solving.

The RLHF stage unexpectedly damages performance: \texttt{GSM8K (Test)} drops sharply to 27.90\%, and \texttt{AP Statistics} remains stagnant. The most plausible explanation is that the Base model lacks sufficient instruction-following capability to interpret preference signals. Without a reliable understanding of prompts or response structure, the model cannot effectively learn the semantic differences between preferred and rejected responses ($y_{w}$ vs. $y_{l}$), resulting in noisy or misdirected optimization.

In summary, Pipeline 1 demonstrates that early knowledge injection without underlying instruction-following capability is ineffective. Domain knowledge alone is insufficient for task-level improvement and is difficult to leverage during subsequent stages.

\subsection{Pipeline 2: Adding instruction tuning to bridge knowledge and task ability}

Pipeline 2 attempts to address the weaknesses of Pipeline 1 by inserting a general-purpose instruction-tuning stage between CoP and SFT. The intent is to strengthen the model's ability to interpret tasks and interact meaningfully with SFT and preference data. Results are shown in Table \ref{tab:pipeline2_results}.

\begin{table}[htbp]
\footnotesize
\centering
\caption{Benchmark results for training Pipeline 2 (knowledge + instruction-bridge strategy).}
\label{tab:pipeline2_results}
\renewcommand{\arraystretch}{1.2}
\begin{threeparttable}
\begin{tabular}{lccc}
\toprule
\textbf{Model} & \textbf{GSM8K (Test)} & \textbf{AP Statistics} & \textbf{ARC} \\
 & (8-shot + CoT, Acc.\ \%) & (0-shot, Acc.\ \%) & (0-shot, Acc.\ \%) \\
\midrule
LLaMA-3.2-3B + CoP                           & 29.49          & \textbf{26.13}          & 42.24 \\
\hline
+ Instruct                           & 26.69          & 25.78          & 42.06 \\
+ Instruct + SFT                     & \textbf{36.47} & 25.78          & \textbf{43.52} \\
+ Instruct + SFT + RLHF              & 26.99          & \textbf{26.13} & 43.17 \\
\bottomrule
\end{tabular}
\begin{tablenotes}
\item Values in boldface denote the best result for each benchmark.
\end{tablenotes}
\end{threeparttable}
\normalsize
\end{table}

After instruction tuning, the performance on all benchmarks decreases slightly relative to the CoP stage. This temporary decline is expected: the model is adapting to a new learning objective—understanding and following instructions—which may disrupt prior patterns. However, this stage is valuable because it equips the model to interpret subsequent SFT data more effectively.

The impact of instruction tuning becomes clear in the SFT stage: \texttt{GSM8K (Test)} improves significantly to 36.47\%, notably higher than the Pipeline 1 counterpart (33.89\%). This confirms the hypothesis: a model that understands instructions learns more effectively from SFT. However, \texttt{AP Statistics} remains unchanged, suggesting that knowledge from CoP does not transfer effectively to domain reasoning tasks, even when the model can now follow instructions.

Similar to Pipeline 1, RLHF degrades performance: \texttt{GSM8K (Test)} falls sharply after RLHF (26.99\%), \texttt{AP Statistics} shows only marginal fluctuations, and \texttt{ARC} declines slightly to 43.17\%. The pattern mirrors Pipeline 1: despite improved instruction-following skills, the model still lacks a deeply integrated knowledge structure. Consequently, preference optimization becomes unstable.

In summary, Pipeline 2 provides only modest improvements over Pipeline 1—mainly on \texttt{GSM8K (Test)}—while still showing no gains in statistical reasoning, weak integration of domain knowledge, and unstable behavior during preference optimization. These results suggest that adding instruction-following capabilities after CoP is considerably less effective than starting with a model that already possesses strong instruction-handling abilities, a limitation that Pipeline 3 is specifically designed to overcome.

\subsection{Pipeline 3: An efficient path for domain adaptation and optimization}

Pipeline 3 constitutes the core experimental direction of this study. Its central strategy is to directly leverage the strong general language understanding and instruction-following capabilities of LLaMA-3.2-3B-Instruct, using it as an efficient starting point for specialization in the statistics domain. Within this pipeline, we designed and rigorously compared three main SFT strategies (v1, v2, and v3.x) to systematically evaluate the combined effects of different fine-tuning methods, data mixing schemes, and preference optimization techniques (GRPO vs. DPO). Finally, we assessed the model's downstream adaptability through a controlled DTFT stage.

\subsubsection{SFT stage: Design and comparison of multi-version strategies (v1, v2, v3.x)}
\label{sss:SFT}

In the SFT stage of Pipeline 3, our primary goal was to steer the general capabilities of the LLaMA-3.2-3B-Instruct model toward statistics-focused reasoning and expression. To that end, we implemented three distinct SFT strategies, each reflecting a different philosophy of training and data integration.

\textbf{SFT-v1 (Phased, differentiated strategy)}: The v1 strategy adopted a multi-phase, template-specific design that emphasized hierarchical capability building:
\begin{itemize}
\item Phase 1: Strengthened mathematical and statistical reasoning through instruction-style SFT on chain-of-thought (\texttt{Statistical CoT}) and mathematical problem QA (\texttt{Math-QA}).
\item Phase 2: Expanded domain expression using \texttt{S2ORC}-derived summary/method/conclusion segments, fine-tuned in an instruction style to enhance statistical narrative fluency.
\item Phase 3: Applied a chat template to separately fine-tune on \texttt{FineTome-100k} (general conversation) and \texttt{Statistical Nouns/Defs} (terminology). LoRA adapter weights from both fine-tunings were then merged via linear interpolation (weighted averaging for the same key values), aiming to integrate domain knowledge, term understanding, and general interaction ability into a single adapter.
\end{itemize}

\textbf{SFT-v2 (Unified, single-stage strategy)}: Reflecting on the complexity and fragility of v1, the v2 strategy pursued a simpler, unified training approach. From the outset, all data were formatted with a chat template, and the following datasets were combined into a single mixed training set: \texttt{Statistical Nouns/Defs}, \texttt{Statistical CoT}, \texttt{Statistical GRPO} (preference-style data used here as supervised signals), \texttt{FineTome-100k}, and \texttt{GSM8K (Train)}. This mixture was used for a one-stage SFT with a higher LoRA rank (rank = 32). The goal was to cultivate comprehensive statistical reasoning and conversational ability within a consistent interaction framework, thereby providing a solid base for subsequent preference optimization.

\textbf{SFT-v3.x (Controlled variants based on v2)}: The v3.x series built upon the unified template design of v2 but introduced controlled variations to examine the effect of different data mixtures and hyperparameters (Table \ref{tab:pipeline3_sft_v3_config}).

\begin{table}[htbp]
\footnotesize
\centering
\caption{Configuration of data mixtures and hyperparameters for the SFT-v3.x series of Pipeline 3.}
\label{tab:pipeline3_sft_v3_config}
\renewcommand{\arraystretch}{1.2}
\begin{threeparttable}
\begin{tabular}{lccccccc}
\toprule
\textbf{Version} & \textbf{Statistical} & \textbf{Statistical} & \textbf{Statistical} & \textbf{FineTome} & \textbf{GSM8K} & \texttt{train\_on} &  \\
& \textbf{Nouns/Defs}\tnote{1} & \textbf{CoT}\tnote{1} & \textbf{GRPO}\tnote{1} & \textbf{-100k}\tnote{1} & \textbf{(Train)}\tnote{1} & \texttt{\_responses\_only}\tnote{2} & \textbf{Epoch}\tnote{3} \\
\midrule
SFT-v3.1 & 3 times & 1 time & 1 time & 1 time & 1 time & $\bigcirc$ & 3 \\
SFT-v3.2 & 3 times & 2 times & 2 times & 1 time & 1 time & $\bigcirc$ & 3 \\
SFT-v3.3 & 3 times & 1 time & 1 time & 1 time & 1 time & $\times$ & 3 \\
SFT-v3.4 & 3 times & 2 times & 2 times & $\times$ & 1 time & $\bigcirc$ & 3 \\
SFT-v3.5 & $\times$ & $\times$ & $\times$ & $\times$ & 1 time & $\bigcirc$ & 1 \\
SFT-v3.6 & 3 times & 1 time & 1 time & $\times$ & $\times$ & $\bigcirc$ & 1 \\
\bottomrule
\end{tabular}
\begin{tablenotes}
\item[1] The entries ``1 time'', ``2 times'', and ``3 times'' indicate the repetition factor for each source dataset when constructing the final mixed training set. A $\times$ symbol in a dataset column indicates that the dataset was not used in that version.
\item[2] \texttt{train\_on\_responses\_only}: $\bigcirc$ indicates loss computation only on the response; $\times$ indicates loss on the full sequence.
\item[3] Number of training epochs.
\end{tablenotes}
\end{threeparttable}
\normalsize
\end{table}

The benchmark results for all SFT variants of Pipeline 3 are reported in Table \ref{tab:pipeline3_sft_results}. From Table \ref{tab:pipeline3_sft_results}, several conclusions emerge:
\begin{enumerate}
\item Domain gains on \texttt{AP Statistics}. All SFT variants improve \texttt{AP Statistics} relative to the original Instruct baseline (37.63\% up to 41.46\%), confirming that SFT is effective at injecting domain-specific statistical knowledge.
\item Trade-offs across benchmarks. SFT-v1 shows modest \texttt{AP Statistics} gains and relatively stable \texttt{ARC}, but suffers a notable drop in \texttt{GSM8K (Test)}, suggesting that its multi-stage complexity does not translate into superior domain performance. SFT-v2 improves \texttt{AP Statistics} more clearly and maintains \texttt{GSM8K (Test)} and \texttt{ARC} at levels comparable to v1, providing a more balanced profile. The v3.x series reveals a nuanced trade-off surface: Configurations like v3.1, v3.3, and v3.4 push \texttt{AP Statistics} higher but tend to reduce \texttt{GSM8K (Test)} and, in some cases, \texttt{ARC} (e.g., v3.3). v3.6 better preserves \texttt{GSM8K (Test)} and \texttt{ARC} but sacrifices some \texttt{AP Statistics} gains.
\item Impact of \texttt{train\_on\_responses\_only} and data composition. Differences between v3.2 vs. v3.3 or v3.4 vs. v3.6 highlight that both the loss computation setting and dataset composition/repetition play important roles in shaping the balance between general and domain-specific abilities.
\end{enumerate}
In summary, the SFT exploration underscores that no single strategy is universally optimal; the best choice depends on the desired trade-off between general reasoning and statistical expertise. SFT-v2 offers a strong, balanced baseline, while variants like v3.3 and v3.4 provide stronger statistical performance at some cost to general abilities. These models (especially v2, v3.3, and v3.4) were selected as candidates for the subsequent preference optimization stage.

\begin{table}[htbp]
\footnotesize
\centering
\caption{Benchmark results of SFT strategies (v1, v2, and v3.x series) in Pipeline 3.}
\label{tab:pipeline3_sft_results}
\renewcommand{\arraystretch}{1.2}
\begin{threeparttable}
\begin{tabular}{lccc}
\toprule
\textbf{Model} & \textbf{GSM8K (Test)} & \textbf{AP Statistics} & \textbf{ARC} \\
 & (8-shot + CoT, Acc.\ \%) & (0-shot, Acc.\ \%) & (0-shot, Acc.\ \%) \\
\midrule
LLaMA-3.2-3B-Instruct & \textbf{64.44} & 37.63 & 43.60 \\
\hline
+ SFT-v1               & 57.92 & 37.98 & 41.89 \\
+ SFT-v2               & 58.53 & 39.72 & 40.36 \\
+ SFT-v3.1             & 51.18 & 41.11 & 40.10 \\
+ SFT-v3.2             & 50.80 & 40.77 & 40.61 \\
+ SFT-v3.3             & 53.60 & \textbf{41.46} & 36.77 \\
+ SFT-v3.4             & 54.59 & 41.11 & 40.61 \\
+ SFT-v3.5             & 55.08 & 39.72 & 39.85 \\
+ SFT-v3.6             & 58.15 & 38.68 & \textbf{41.98} \\
\bottomrule
\end{tabular}
\begin{tablenotes}
\item Values in boldface denote the best result for each benchmark.
\end{tablenotes}
\end{threeparttable}
\normalsize
\end{table}

\subsubsection{RLHF preference optimization: From GRPO limitations to effective DPO alignment}
\label{sss:RLHF}

After obtaining a set of SFT models with different capability profiles, we moved to the RLHF preference optimization stage, aiming to use preference-based supervision to improve output quality, coherence, and alignment with human expectations beyond what standard SFT can achieve.

\textbf{GRPO experiments}: As an initial attempt along the RLHF-style preference optimization path, we applied GRPO to the SFT-v1 and SFT-v2 models. For SFT-v2, we further explored three GRPO configurations by varying: LoRA rank and alpha, and the data mixing ratio between \texttt{Statistical GRPO} and \texttt{GSM8K (Train)} samples. Table \ref{tab:pipeline3_rlhf_grpo_config} summarizes the GRPO settings, and Table \ref{tab:pipeline3_rlhf_grpo_results} reports the benchmark scores.

\begin{table}[htbp]
\footnotesize
\centering
\caption{GRPO hyperparameter configurations used with SFT-v1 and SFT-v2 in Pipeline 3.}
\label{tab:pipeline3_rlhf_grpo_config}
\renewcommand{\arraystretch}{1.2}
\begin{threeparttable}
\begin{tabular}{lcc}
\toprule
\textbf{Version} & \textbf{LoRA Rank} & \textbf{Data Mixing Ratio}\tnote{1} \\
& \textbf{\& Alpha} & (\texttt{Statistical GRPO} : \texttt{GSM8K (Train)}) \\
\midrule
GRPO-v1 & 32 & 2255 : 4000 \\
GRPO-v2 & 8  & 2255 : 1000 \\
GRPO-v3 & 16 & 2255 : 1000 \\
\bottomrule
\end{tabular}
\begin{tablenotes}
\item[1] The sample ratio between \texttt{Statistical GRPO} data and \texttt{GSM8K (Train)} data used for training.
\end{tablenotes}
\end{threeparttable}
\normalsize
\end{table}

\begin{table}[htbp]
\footnotesize
\centering
\caption{Benchmark results from the GRPO preference optimization stage used with SFT-v1 and SFT-v2 in Pipeline 3.}
\label{tab:pipeline3_rlhf_grpo_results}
\renewcommand{\arraystretch}{1.2}
\begin{threeparttable}
\begin{tabular}{lccc}
\toprule
\textbf{Model} & \textbf{GSM8K (Test)} & \textbf{AP Statistics} & \textbf{ARC} \\
 & (8-shot + CoT, Acc.\ \%) & (0-shot, Acc.\ \%) & (0-shot, Acc.\ \%) \\
\midrule
LLaMA-3.2-3B-Instruct & \textbf{64.44} & 37.63 & \textbf{43.60} \\
\hline
+ SFT-v1              & 57.92 & 37.98 & 41.89 \\
+ SFT-v1 + GRPO-v2    & 58.45 & 38.33 & 42.32 \\
\hline
+ SFT-v2              & 58.53 & 39.72 & 40.36 \\
+ SFT-v2 + GRPO-v1    & 55.42 & 40.07 & 39.85 \\
+ SFT-v2 + GRPO-v2    & 53.98 & \textbf{40.42} & 39.85 \\
+ SFT-v2 + GRPO-v3    & 57.16 & 40.07 & 35.32 \\
\bottomrule
\end{tabular}
\begin{tablenotes}
\item Values in boldface denote the best result for each benchmark.
\end{tablenotes}
\end{threeparttable}
\normalsize
\end{table}

The GRPO results reveal several issues. For SFT-v1, GRPO-v2 yields only minor gains in \texttt{GSM8K (Test)} and \texttt{ARC}, and almost no meaningful improvement in \texttt{AP Statistics}, suggesting limited domain enhancement. For SFT-v2, the behavior is strongly dependent on hyperparameters: GRPO-v1 (highest rank, alpha, and largest \texttt{GSM8K (Train)} share) slightly increases \texttt{AP Statistics} but significantly reduces \texttt{GSM8K (Test)}. GRPO-v2 (lowest rank, alpha, and less \texttt{GSM8K (Train)}) achieves the highest \texttt{AP Statistics} but at the cost of a substantial drop in \texttt{GSM8K (Test)}. GRPO-v3 (intermediate settings) partially recovers \texttt{GSM8K (Test)} but fails to surpass GRPO-v2 on \texttt{AP Statistics} and causes a severe degradation on \texttt{ARC}, approaching a catastrophic drop.

Overall, GRPO proved highly sensitive to hyperparameter choices and data mixing ratios, often redistributing performance rather than reliably improving it. In our setting, it did not provide stable, consistent, or controllable gains and frequently harmed general abilities in exchange for modest domain improvements. Given its instability and complexity, we concluded that GRPO was not a suitable primary alignment method for this study.

\textbf{DPO Experiments}: We therefore shifted our focus to DPO, which reframes preference learning as a direct classification-style optimization on paired preference data, avoiding explicit reward modeling and complex reinforcement learning updates. Theoretically, this offers a more stable and straightforward training process.

We applied DPO to three key SFT models: SFT-v2 (balanced capabilities), SFT-v3.3 and SFT-v3.4 (stronger AP Statistics but weakened general performance). The DPO results are summarized in Table \ref{tab:pipeline3_rlhf_dpo_results}.

\begin{table}[htbp]
\footnotesize
\centering
\caption{Benchmark results from the DPO preference optimization stage used with SFT-v2 and SFT-v3.x in Pipeline 3.}
\label{tab:pipeline3_rlhf_dpo_results}
\renewcommand{\arraystretch}{1.2}
\begin{threeparttable}
\begin{tabular}{lccc}
\toprule
\textbf{Model} & \textbf{GSM8K (Test)} & \textbf{AP Statistics} & \textbf{ARC} \\
 & (8-shot + CoT, Acc.\ \%) & (0-shot, Acc.\ \%) & (0-shot, Acc.\ \%) \\
\midrule
LLaMA-3.2-3B-Instruct & \textbf{64.44} & 37.63 & \textbf{43.60} \\
\hline
+ SFT-v2            & 58.53 & 39.72 & 40.36 \\
+ SFT-v2 + DPO      & 57.85 & 40.07 & 40.27 \\
\hline
+ SFT-v3.3          & 53.60 & \textbf{41.46} & 36.77 \\
+ SFT-v3.3 + DPO    & 57.24 & \textbf{41.46} & 41.13 \\
\hline
+ SFT-v3.4          & 54.59 & 41.11 & 40.61 \\
+ SFT-v3.4 + DPO    & 58.98 & \textbf{41.46} & 41.81 \\
\bottomrule
\end{tabular}
\begin{tablenotes}
\item Values in boldface denote the best result for each benchmark.
\end{tablenotes}
\end{threeparttable}
\normalsize
\end{table}

DPO exhibits a much more favorable profile. For SFT-v3.3 and SFT-v3.4, DPO substantially restores general abilities (\texttt{GSM8K (Test)}, \texttt{ARC}) that had been weakened during SFT while preserving or slightly improving \texttt{AP Statistics}. SFT-v3.4 + DPO provides the strongest overall performance, with improvements across \texttt{GSM8K (Test)} and \texttt{ARC} and a slight \texttt{AP Statistics} gain. For SFT-v2, DPO maintains \texttt{AP Statistics} and \texttt{ARC} roughly at their original level while slightly reducing \texttt{GSM8K (Test)}, yielding a relatively balanced but not dominant configuration compared to SFT-v3.4 + DPO.

These results highlight DPO's balancing power: it can restore or improve general reasoning while maintaining domain expertise, rather than trading one off against the other. Among all configurations, SFT-v3.4 + DPO emerges as the most balanced and highest-performing model across the three benchmarks and becomes the primary candidate for the final downstream fine-tuning stage.

\subsubsection{Downstream task fine-tuning: Effects of different fine-tuning intensities}
\label{sss:DTFT}

Having obtained SFT-v3.4 + DPO as a strong and well-balanced statistical LLM, Pipeline 3 proceeds to its final stage: DTFT. This stage serves two core purposes:
\begin{enumerate}
\item Task adaptation: Further refine performance on high-quality, domain-specific QA datasets—\texttt{Cross Validated} (\texttt{CV}) and \texttt{Knowledge Graph} (\texttt{KG})—without damaging the statistical knowledge and general reasoning abilities established in earlier stages.
\item Conversational capability exploration: Investigate the impact of adding \texttt{Statistical Conversation} (\texttt{SC}) data, given that the model is already based on an Instruct backbone and trained with a chat template.
\end{enumerate}
The main challenge is to adjust the model to downstream tasks without disrupting the finely tuned balance of abilities. To study this, we designed five DTFT configurations with different intensities and data combinations, summarized in Table \ref{tab:pipeline3_dtft_config}.

\begin{table}[htbp]
\footnotesize
\centering
\caption{Configuration details of different strategies in the DTFT stage of Pipeline 3.}
\label{tab:pipeline3_dtft_config}
\renewcommand{\arraystretch}{1.2}
\begin{tabular}{lcccc}
\toprule
\textbf{Version} & \textbf{LoRA Rank} & \textbf{LoRA Alpha} & \textbf{Training Duration} & \textbf{Training Data} \\
 & & & & \textbf{Combination} \\
\midrule
DTFT-v1 & 16 & 32 & 1 epoch    & \texttt{CV}, \texttt{KG} \\
DTFT-v2 & 8  & 16 & 150 steps  & \texttt{CV}, \texttt{KG} \\
DTFT-v3 & 8  & 16 & 300 steps  & \texttt{CV}, \texttt{KG} \\
DTFT-v4 & 8  & 16 & 180 steps  & \texttt{CV}, \texttt{KG}, \texttt{SC} \\
DTFT-v5 & 8  & 16 & 300 steps  & \texttt{CV}, \texttt{KG}, \texttt{SC} \\
\bottomrule
\end{tabular}
\normalsize
\end{table}

The benchmark results for these configurations are shown in Table \ref{tab:pipeline3_dtft_results}. Several key observations arise:
\begin{enumerate}
\item Sensitivity to fine-tuning intensity. DTFT-v1, which uses a full epoch of fine-tuning with \texttt{CV} and \texttt{KG}, causes performance to drop across all three benchmarks. This underscores that even high-quality, domain-relevant data can disrupt a finely balanced model when fine-tuned too aggressively, leading to overfitting and potential catastrophic forgetting. This confirms that for highly optimized models, conventional full-epoch fine-tuning is often too strong and must be replaced with more carefully controlled strategies.
\item Effectiveness of low-intensity fine-tuning. DTFT-v2, with a very conservative training duration (150 steps), almost perfectly preserves the peak \texttt{GSM8K (Test)} and \texttt{AP Statistics} performance of SFT-v3.4 + DPO, with only a minor reduction on \texttt{ARC}. This demonstrates that carefully constrained fine-tuning can successfully adapt the model to downstream data without degrading its core capabilities.
\item Non-linear effects of slightly increased intensity. DTFT-v3 doubles the steps (300) with the same data as DTFT-v2. While \texttt{AP Statistics} is preserved and \texttt{ARC} improves slightly, \texttt{GSM8K (Test)} decreases. This indicates non-linear sensitivity: even within a low-intensity regime, small changes in training duration can shift the balance among capabilities.
\item Impact of adding conversational data (\texttt{SC}). DTFT-v4 and DTFT-v5, which incorporate \texttt{SC} data along with \texttt{CV} and \texttt{KG}, show mixed results. Relative to DTFT-v2, \texttt{GSM8K (Test)} drops slightly in both v4 and v5; \texttt{AP Statistics} decreases in v4 but is restored in v5 to the level of v2; and \texttt{ARC} improves modestly in both v4 and v5. These patterns suggest that adding conversational data does not uniformly improve benchmark performance, and may introduce noise or shifts in style that slightly interfere with existing abilities.
\end{enumerate}
Overall, these results highlight that downstream fine-tuning must be applied with extreme care in the final stages of a multi-stage training pipeline. Slight changes in training intensity or data composition can materially affect the balance between general reasoning and domain expertise.

\begin{table}[htbp]
\footnotesize
\centering
\caption{Benchmark results from the DTFT stage, starting from SFT-v3.4 + DPO in Pipeline 3.}
\label{tab:pipeline3_dtft_results}
\renewcommand{\arraystretch}{1.2}
\begin{threeparttable}
\begin{tabular}{lccc}
\toprule
\textbf{Model} & \textbf{GSM8K (Test)} & \textbf{AP Statistics} & \textbf{ARC} \\
 & (8-shot + CoT, Acc.\ \%) & (0-shot, Acc.\ \%) & (0-shot, Acc.\ \%) \\
\midrule
LLaMA-3.2-3B-Instruct & \textbf{64.44} & 37.63 & \textbf{43.60} \\
+ SFT-v3.4 + DPO                    & 58.98 & \textbf{41.46} & 41.81 \\
\hline
+ DTFT-v1                        & 56.25 & 41.11 & 40.27 \\
+ DTFT-v2                        & 58.83 & \textbf{41.46} & 40.61 \\
+ DTFT-v3                        & 57.39 & \textbf{41.46} & 41.13 \\
+ DTFT-v4                        & 58.38 & 40.77 & 40.70 \\
+ DTFT-v5                        & 58.30 & \textbf{41.46} & 40.78 \\
\bottomrule
\end{tabular}
\begin{tablenotes}
\item Values in boldface denote the best result for each benchmark.
\end{tablenotes}
\end{threeparttable}
\normalsize
\end{table}

Combining insights from the SFT, RLHF, and DTFT stages, we find that SFT-v3.4 + DPO under the DTFT-v2 configuration offers the best overall trade-off: it preserves strong general reasoning, maintains high \texttt{AP Statistics} performance, and remains robust on \texttt{ARC}, while successfully adapting to high-quality, domain-specific QA data. We designate this final model as StatLLaMA. In the subsequent qualitative analysis, we focus on examining StatLLaMA’s behavior and responses in depth.

\subsection{Qualitative analysis: In-depth examination of model capabilities}

In addition to the quantitative benchmarks, we conducted a qualitative assessment to further examine the statistical proficiency and interactive behavior of the proposed model. This analysis focuses on the extent to which the final model, StatLLaMA, exhibits improved conceptual accuracy, reasoning depth, and adherence to task constraints relative to its initialization model, LLaMA-3.2-3B-Instruct.

We evaluated three representative use cases: (i) standard conceptual question answering, (ii) knowledge-graph–based factual querying grounded in the STATO/OBCS ontological structure, and (iii) multi-turn consultation scenarios that approximate applied statistical reasoning. For each case, we constructed prompts targeting domain-relevant constructs, including foundational inferential concepts, ontology-based entity relations, and practical data-analytic decision making. Responses were independently reviewed by researchers with statistical training using the following criteria: conceptual correctness, logical coherence, clarity of exposition, terminological precision, mitigation of common misconceptions, and appropriateness of interaction flow. Representative examples are provided in Appendix Table \ref{atab:model_samples}.

Across the three scenarios, StatLLaMA demonstrated notably improved performance. In the conceptual question-answering task (Appendix Table \ref{atab:model_samples}, Prompt 1), its explanation of the distinction between descriptive and inferential statistics was more structured and aligned with standard pedagogical formulations, and the example invoking confidence intervals accurately illustrated inferential reasoning. In the knowledge-graph query task (Appendix Table \ref{atab:model_samples}, Prompt 2), StatLLaMA adhered more closely to the requested schema, producing responses that remained within the specified ontology and avoided extraneous information—an error exhibited by the baseline model. In the consulting scenario (Appendix Table \ref{atab:model_samples}, Prompt 3), StatLLaMA provided recommendations consistent with accepted practice, including identifying repeated-measures ANOVA as an appropriate method, articulating relevant assumptions such as sphericity, and suggesting mixed-effects modeling as an alternative. These responses indicate a higher level of procedural and conceptual integration.

Taken together, the qualitative evidence complements the quantitative results reported earlier. StatLLaMA exhibits improved domain alignment, more coherent reasoning, and more context-appropriate interaction behavior. The examples in Appendix Table \ref{atab:model_samples} illustrate how multi-stage training introduced in this study leads to measurable gains in applied statistical competence.

\section{Conclusion}

\subsection{Summary of findings}

This study systematically examined multi-stage training strategies for developing a statistical LLM capable of integrating domain-specific statistical knowledge with strong general-purpose reasoning abilities. Using the lightweight LLaMA-3.2-3B family as the foundation, our objective was to construct a model deplorable in resource-constrained environments while achieving high accuracy on tasks requiring statistical expertise.

By empirically comparing three representative training pipelines, we demonstrate that the choice of starting model critically determines the efficiency and success of domain adaptation. Pipelines that begin from the base model—without pre-existing instruction-following capability—struggle to absorb domain knowledge, even after extensive CoP and subsequent instruction tuning. As a result, performance on core statistical reasoning tasks remains limited, and later SFT or preference-alignment stages often yield marginal or even negative gains.

In contrast, Pipelines 3, which begins with LLaMA-3.2-3B-Instruct, exhibits substantially better training efficiency. Starting from a model with strong instruction-following ability allows subsequent SFT to more effectively specialize the model toward statistical reasoning. However, our SFT experiments (v1, v2, v3.x) highlight a clear trade-off: pushing domain specialization too far can degrade general reasoning performance on benchmarks such as \texttt{GSM8K (Test)} and \texttt{ARC}. The v3.x series systematically quantified this balance and enabled principled selection of model variants.

In the alignment stage, DPO consistently outperformed GRPO. GRPO demonstrated high sensitivity to hyperparameters and produced unstable capability trade-offs across reasoning metrics. DPO, by contrast, improved output quality reliably, aligned responses more closely with disciplinary norms in statistics, and—importantly—restored general abilities diminished during SFT. The best-performing model from the SFT stage (v3.4) improved further under DPO, producing the most balanced and robust model in the study.

Finally, our analysis of DTFT revealed that when a model has undergone substantial prior optimization, only extremely low-intensity fine-tuning is viable. Conventional or moderate-intensity fine-tuning quickly destabilizes earlier gains and induces substantial performance degradation. A carefully calibrated low-step, low-rank LoRA configuration (DTFT-v2) successfully adapted the model to high-quality domain-specific QA data while preserving core capabilities. Even slight increases in training intensity led to disproportionate declines. Additional experiments incorporating statistical conversation data produced mixed benefits, suggesting the importance of careful alignment between data type and evaluation objectives.

Based on these results, the combination LLaMA-3.2-3B-Instruct + SFT-v3.4 + DPO + DTFT-v2 was identified as optimal. This final model—StatLLaMA—achieved strong and stable performance across \texttt{AP Statistics}, \texttt{GSM8K (Test)}, and \texttt{ARC}, effectively integrating statistical expertise with general linguistic reasoning. More broadly, the study establishes an empirically grounded workflow for constructing domain-specialized yet lightweight LLMs:

High-quality Instruct initialization $\rightarrow$ Domain-oriented SFT $\rightarrow$ Stable DPO preference alignment  $\rightarrow$ Minimal, controlled DTFT.

This framework offers a practical blueprint for future development of professional-domain LLMs.

\subsection{Limitations}

Despite the contributions of this study, several limitations warrant consideration.

First, model scale imposes a clear constraint. All experiments were conducted on the 3B parameter LLaMA-3.2 models. While this size is suitable for resource-limited deployment, its capacity is markedly lower than that of 7B, 70B, or larger models. It remains unclear whether the learning dynamics, trade-offs, and optimal strategies identified here would generalize to larger architectures, which may exhibit different sensitivity to CoP, SFT, or alignment signals.

Second, the evaluation suite, though diverse, remains incomplete. \texttt{GSM8K (Test)}, \texttt{ARC}, and \texttt{AP Statistics} provided coverage of mathematical reasoning, common-sense reasoning, and core statistical knowledge. Yet they do not fully assess real-world statistical practice, including handling noisy datasets, reasoning under ambiguity, performing multistep workflows (data cleaning $\rightarrow$ modeling $\rightarrow$ interpretation), or assessing misuse of methods in applied contexts. These areas will require more comprehensive benchmarks.

Third, despite careful data curation, topic coverage remains imperfect. Statistics is broad, and certain subfields may be underrepresented in training data. Additionally, some preference data was distilled from the Gemini API. While effective for performance, this may introduce teacher-model biases and reduce epistemic diversity. A shortage of high-quality statistical chain-of-thought data also limits the model's potential to improve interpretability and reasoning transparency.

Fourth, the DTFT stage, though informative, remains preliminary. While low-intensity fine-tuning proved essential, we did not exhaustively explore the space of PEFT configurations, hyperparameters, or data compositions. Our investigation into statistical conversation data was also incomplete, leaving open questions regarding optimal integration strategies and effects on real-world conversational performance.

Finally, resource constraints limited the number of independent experimental runs and prevented comprehensive hyperparameter sweeps. Although the main conclusions are supported by consistent trends, stochastic variability and unexplored parameter combinations may conceal additional optimal strategies or alternative explanations.

\subsection{Future directions}

This study reveals several promising directions for future research.

Scaling the pipeline to larger models will allow examination of whether the training dynamics and strategy effectiveness generalize beyond the 3B scale. Larger models may internalize domain knowledge more efficiently or exhibit more stable behavior in alignment and fine-tuning.

Developing more comprehensive evaluation frameworks is essential. Future benchmarks should include end-to-end applied statistical tasks and more sophisticated conversational evaluations, including ambiguity resolution, error diagnosis, and critique of flawed analyses.

Improving data quality and breadth remains a major opportunity. Future work should expand high-quality statistical reasoning datasets, construct richer CoT corpora, and investigate improved distillation strategies that reduce teacher-model bias while maintaining signal strength. Methods for structured data augmentation, counterexample generation, and topic balancing warrant exploration.

Advances in fine-tuning and PEFT techniques also hold potential. Comparing LoRA variants, exploring AdapterFusion \citep{Pfeifferetal2021adapterfusion}, or partially freezing modules may produce better stability–adaptability trade-offs.

Finally, explainability, robustness, and safety will become increasingly important as statistical LLMs are deployed in high-stakes domains. Future research should examine internal reasoning mechanisms, bias and failure modes, adversarial robustness, and uncertainty quantification to ensure reliability in practical use cases.

Overall, this study establishes a strong foundation for the development of statistical LLMs. As training algorithms, evaluation protocols, and data resources continue to advance, the potential applications of specialized LLMs in statistical analysis, decision support, and education will expand significantly.

\section*{Acknowledgments}

This research was partially supported by grants from the Ministry of Science and Technology, Taiwan (MOST 111-2118-M-A49-003-MY2) and the National Science and Technology Council, Taiwan (NSTC 113-2118-M-A49-006 and NSTC 114-2118-M-A49-002). We are grateful to the National Center for High-performance Computing, Taiwan for computer time and facilities. During the preparation of this work, the authors used OpenAI's ChatGPT in order to improve language and readability. After using this tool, the authors reviewed and edited the content as needed and take full responsibility for the content of the publication.

\bibliographystyle{IEEETranN}
\bibliography{StatLLaMA_arxiv}

\newpage

\section*{Appendix}


\footnotesize
\begin{longtable}[l]{p{8cm} p{8cm}}
\caption{Prompts used to evaluate statistical question-answering performance, along with model outputs from StatLLaMA and LLaMA-3.2-3B-Instruct.}
\label{atab:model_samples} \\
\hline
\endfirsthead

\multicolumn{2}{c}%
{Table \thetable\ -- Continued from previous page} \\
\hline
\endhead

\multicolumn{2}{r}{{Continued on next page}} \\
\endfoot

\hline
\endlastfoot

\multicolumn{2}{l}{\textbf{Prompt 1:}} \\
\multicolumn{2}{l}{\textit{Explain the difference between descriptive statistics and inferential statistics. }} \\
\multicolumn{2}{l}{\textit{Provide an example of each.}} \\

\hline

\textbf{LLaMA-3.2-3B-Instruct:}
\vspace{0.3cm}

Descriptive statistics and inferential statistics are two types of statistical methods used to analyze and interpret data. The main difference between them lies in their purpose, scope, and level of inference.

\textbf{Descriptive Statistics}

Descriptive statistics aim to summarize and describe the main features of a dataset. They provide a snapshot of the data, without making assumptions about the population. The main goals are:
\begin{enumerate}
\item To summarize the central tendency (mean, median, mode)
\item To summarize variability (range, variance, standard deviation)
\item To describe shape and distribution (skewness, kurtosis)
\end{enumerate}

\textit{Example:} A researcher describes the average height of 100 students by calculating the mean (175 cm). This summarizes the sample without making population-level inferences.

\textbf{Inferential Statistics}

Inferential statistics aim to make inferences about a population based on a sample. The main goals are:
\begin{enumerate}
\item To estimate population parameters (mean, proportion)
\item To test hypotheses (e.g., is the population mean equal to a known value?)
\item To make predictions about the population
\end{enumerate}

\textit{Example:} A researcher collects a sample of 100 adults to test if the population mean height exceeds 175 cm. If the sample mean is greater, the researcher infers that the population mean is likely greater as well.

\textbf{Summary}

Descriptive statistics summarize data; inferential statistics draw conclusions about populations from samples. &

\textbf{StatLLaMA:}
\vspace{0.3cm}

Descriptive statistics and inferential statistics are two major types of methods used to analyze data.

\textbf{Descriptive Statistics}

Descriptive statistics summarize and describe the basic features of a dataset, helping to understand characteristics such as central tendency, variability, and distribution. They provide a snapshot of the data without inferring population properties.

\textit{Example:} Given exam scores from 100 students, we can compute the mean, median, mode, and range to describe the class’s performance distribution.

\textbf{Inferential Statistics}

Inferential statistics are used to make generalizations about a population based on a sample. They allow conclusions about characteristics, relationships, or behaviors of the population.

\textit{Example:} To determine whether the average exam score in one school is higher than the national average, we compute a confidence interval for the sample mean. If this interval exceeds the national average, we infer the school’s average is likely higher.

\textbf{Key Differences}
\begin{itemize}
\item Descriptive statistics summarize observed data, while inferential statistics make predictions about the population.
\item Descriptive methods describe characteristics; inferential methods generalize findings.
\item Descriptive analysis does not involve inference, whereas inferential analysis uses probability to draw conclusions.
\end{itemize} \\

\hline

\newpage

\multicolumn{2}{l}{\textbf{Prompt 2:}} \\
\multicolumn{2}{l}{\textit{What are some key properties or assumptions associated with ``Linear Regression Model'' as defined in the knowledge graph? }} \\
\multicolumn{2}{l}{\textit{Based on triples like:}} \\
\multicolumn{2}{l}{\textit{(Linear Regression Model, has\_assumption, linearity),}} \\
\multicolumn{2}{l}{\textit{(Linear Regression Model, has\_assumption, independence of errors),}} \\
\multicolumn{2}{l}{\textit{(Linear Regression Model, has\_assumption, homoscedasticity),}} \\
\multicolumn{2}{l}{\textit{(Linear Regression Model, has\_assumption, normality of errors)}} \\

\hline

\textbf{LLaMA-3.2-3B-Instruct:}
\vspace{0.3cm}

Based on the provided triples, here are some key properties or assumptions associated with a Linear Regression Model:
\begin{enumerate}
\item \textbf{Linearity}: The relationship between the independent variables and the dependent variable is assumed to be linear. (e.g., (Linear Regression Model, has\_assumption, linearity))
\item \textbf{Independence of Errors}: The errors (residuals) are assumed to be independent of each other. (e.g., (Linear Regression Model, has\_assumption, independence of errors))
\item \textbf{Homoscedasticity}: The variance of the errors is assumed to be constant across all levels of the independent variables. (e.g., (Linear Regression Model, has\_assumption, homoscedasticity))
\item \textbf{Normality of Errors}: The errors are assumed to be normally distributed. (e.g., (Linear Regression Model, has\_assumption, normality of errors))
\item \textbf{Additivity}: The relationship between the independent variables and the dependent variable is assumed to be additive, i.e., the effect of one independent variable is independent of the effects of other independent variables. (not explicitly mentioned in the provided triples, but a common assumption in linear regression)
\item \textbf{No multicollinearity}: The independent variables are assumed to be uncorrelated with each other. (not explicitly mentioned in the provided triples, but a common assumption in linear regression)
\item \textbf{No omitted variables}: The model assumes that all relevant variables are included in the model. (not explicitly mentioned in the provided triples, but a common assumption in linear regression)
\end{enumerate}

These assumptions are essential for the validity and reliability of linear regression models. However, it's worth noting that in practice, it's often difficult to meet all these assumptions simultaneously, and linear regression models may need to be adjusted or modified to account for any violations. &

\textbf{StatLLaMA:}
\vspace{0.3cm}

Based on the knowledge graph, the key properties or assumptions associated with a Linear Regression Model are:
\begin{enumerate}
\item \textbf{Linearity}: The relationship between the independent variables and the dependent variable is assumed to be linear.
\item \textbf{Independence of Errors}: The errors in the model are assumed to be independent of each other.
\item \textbf{Homoscedasticity}: The variance of the errors is assumed to be constant across all levels of the independent variables.
\item \textbf{Normality of Errors}: The errors are assumed to be normally distributed.
\end{enumerate}

These assumptions are fundamental to the Linear Regression Model and are used to estimate the coefficients of the model. Violations of these assumptions can lead to biased or inconsistent estimates, and may require additional steps to account for the violations. \\

\hline

\newpage

\multicolumn{2}{l}{\textbf{Prompt 3:}} \\
\multicolumn{2}{l}{\textit{Hi, I’m a medical intern analyzing patient blood pressure data collected across different times of day.}} \\
\multicolumn{2}{l}{\textit{My supervisor asked me to check whether time of measurement significantly affects blood pressure.}} \\
\multicolumn{2}{l}{\textit{I thought of using ANOVA, but I’m confused because the same patients appear in multiple time slots.}} \\
\multicolumn{2}{l}{\textit{Should I be using repeated measures ANOVA instead?}} \\

\hline

\textbf{LLaMA-3.2-3B-Instruct:}
\vspace{0.3cm}

You're correct to consider the repeated measures aspect of your data. Since the same patients appear in multiple time slots, you're dealing with a type of repeated measures data.

ANOVA (Analysis of Variance) is typically used when you have independent groups being compared to each other, whereas in your case, you have multiple measurements from the same group (patients) at different time points.

Repeated Measures ANOVA (RM-ANOVA) is the appropriate statistical test to use in your situation. RM-ANOVA compares the means of multiple measurements from the same group to detect significant differences between the measurements.

Here's a brief summary of the key differences:
\begin{itemize}
\item \textbf{ANOVA}: Compares means of independent groups (e.g., different patients, treatments, or conditions) to detect significant differences between the groups.
\item \textbf{Repeated Measures ANOVA (RM-ANOVA)}: Compares means of multiple measurements from the same group (e.g., different time points, patients, or conditions) to detect significant differences between the measurements.
\end{itemize}

When using RM-ANOVA, you'll need to consider the following:
\begin{itemize}
\item \textbf{Repeatability}: You need to ensure that the measurements are repeatable, meaning that the same patient's blood pressure is measured at the same time on multiple occasions.
\item \textbf{Randomization}: You should randomize the order of measurements to minimize any biases.
\item \textbf{Outliers}: You'll need to handle any outliers that may affect the results.
\end{itemize}

By using RM-ANOVA, you'll be able to determine whether the time of measurement significantly affects blood pressure in your patients.

It's always a good idea to consult with your supervisor or a statistician for further guidance on your specific data and research question. &

\textbf{StatLLaMA:}
\vspace{0.3cm}

In your case, since the same patients appear in multiple time slots, you're dealing with a repeated measures design. This means that you're essentially measuring the same variable (blood pressure) at different times (time slots) for the same subjects.

Using repeated measures ANOVA is the correct approach here. Here's why:
\begin{enumerate}
\item \textbf{Subject variability}: Since the same patients are measured multiple times, subject variability is a concern. Repeated measures ANOVA accounts for this variability by including it as a random effect in the model.
\item \textbf{Dependent variable}: Blood pressure is a continuous dependent variable, which is suitable for repeated measures ANOVA.
\item \textbf{Time slots}: You have multiple time slots, which are the levels of the repeated measure. Repeated measures ANOVA allows you to compare the means of blood pressure across these time slots.
\end{enumerate}

ANOVA, on the other hand, is typically used for independent samples (e.g., comparing two groups) and doesn't account for subject variability.

When using repeated measures ANOVA, keep in mind:
\begin{enumerate}
\item \textbf{Assumptions}: Check the assumptions of repeated measures ANOVA, such as normality of the data and equal variances across time slots.
\item \textbf{Model complexity}: Consider whether additional models, such as mixed-effects models or generalized linear mixed models, might be more suitable for your data.
\end{enumerate}

In your case, using repeated measures ANOVA should help you determine whether time of measurement significantly affects blood pressure. \\

\end{longtable}
\normalsize

\newpage

\footnotesize
\begin{longtable}[l]{p{3cm} p{13cm}}
\caption{Full set of prompt templates used in data preparation. Each dataset is generated by Gemini using its corresponding prompt, encompassing diverse statistical reasoning and dialogue tasks to ensure comprehensive understanding across training strategies.}
\label{atab:prompt} \\
\hline
\textbf{Dataset} & \textbf{Prompt} \\
\hline
\endfirsthead

\multicolumn{2}{c}%
{Table \ref{atab:prompt} -- Continued from previous page} \\
\hline
\textbf{Dataset} & \textbf{Prompt} \\
\hline
\endhead

\multicolumn{2}{r}{{Continued on next page}} \\
\endfoot

\hline
\endlastfoot

Statistical CoT &
You are a distinguished Statistics Professor and an expert in educational design. Your task is to generate \{num\_to\_generate\} high-quality statistical problem sets. Each problem must include a clear, step-by-step \textit{chain-of-thought} reasoning process. Accuracy, clarity, and pedagogical rigor are essential.

\textbf{Instructions:}

1. \textbf{Output Format:}
Produce a single JSON list containing \{num\_to\_generate\} dictionary objects. Do not include markdown indicators such as \texttt{"json"}. Each object must contain the keys \texttt{"question"} and \texttt{"cot"}.
Example: \texttt{\{"question": "...", "cot": ["1. ...", "2. ...", "3. ..."]\}}.

2. \textbf{Content Guidelines:}
- \texttt{"question"}: A clear, multi-step statistical question covering topics such as probability theory (Bayes’ theorem, conditional probability), distributions (Binomial, Poisson, Normal), sampling, confidence intervals, hypothesis testing (t-tests, chi-squared tests), correlation, and regression.
- \texttt{"cot"}: A list of sequential reasoning steps written as strings. Each step should progress logically toward the final answer. All formulas and variables must be written in valid LaTeX notation. \\

\hline

Statistical GRPO &
You are a distinguished Statistics Professor known for creating clear and instructive problems. Your task is to generate \{num\_to\_generate\} high-quality Question–Answer (QA) sets for Group Relative Policy Optimization (GRPO).

\textbf{Instructions:}

1. \textbf{Output Format:}
Create a single JSON list of dictionaries with keys \texttt{"question"}, \texttt{"reasoning"}, and \texttt{"answer"}. All numerical values must be represented as strings.
Example: \texttt{\{"question": "...", "reasoning": "...", "answer": "0.375"\}}.

2. \textbf{Content Requirements:}
- \texttt{"question"}: Pose a clear statistical question requiring analytical reasoning and multi-step problem solving. Cover diverse areas such as descriptive statistics, probability, distributions, sampling, confidence intervals, hypothesis testing, correlation, and regression. Include both conceptual and computational tasks, avoiding overly simplistic textbook examples.
- \texttt{"reasoning"}: Provide a detailed, accurate explanation with all formulas in LaTeX notation.
- \texttt{"answer"}: Present the definitive answer derived from the reasoning. Ensure consistency with any format requirements specified in the question. \\

\hline

Statistical DPO &
You are an AI assistant tasked with generating training data for Direct Preference Optimization (DPO). Produce \{num\_to\_generate\} distinct data points encompassing various statistical concepts within a single response.

\textbf{Instructions:}

1. \textbf{Output Format:}
Generate one valid JSON list where each element contains the keys \texttt{"prompt"}, \texttt{"chosen"}, and \texttt{"rejected"}.
Example: \texttt{[\{"prompt": "...", "chosen": "...", "rejected": "..." \}, ...]}.

2. \textbf{Content Guidelines:}
- \textbf{Topic Diversity:} Cover different statistical domains, including descriptive statistics, probability, inference, regression, and study design. Aim for a mix of difficulty levels.
- \texttt{"prompt"}: Vary prompt styles — conceptual explanations, comparisons, applications, or reasoning-based tasks.
- \texttt{"chosen"}: Provide accurate, clear, and well-structured answers with sound logic. Use good examples/
analogies where appropriate.
- \texttt{"rejected"}: Address the same prompt but include subtle inaccuracies, misconceptions, or unclear reasoning to create a plausible yet inferior response. It should not be completely wrong or nonsensical.\\

\hline

Statistical Conversation &
You are a knowledgeable AI specializing in statistics education. Generate a collection of realistic, diverse, multi-turn conversations between a “user” and a “statistical assistant.” Each conversation should address a distinct statistical concept, method, or common source of confusion. Create \{num\_to\_generate\} conversations per batch.

\textbf{Instructions:}

1. \textbf{Topic Diversity:}
Cover a broad range of topics including descriptive statistics, probability, inference, hypothesis testing, regression, experimental design, Bayesian reasoning, and common misconceptions.

2. \textbf{Conversation Structure:}
Each conversation should have at least \{min\_turns\} turns (1 user + 1 assistant = 1 turn), ideally \{target\_turns\}. Start with a natural user question, progress logically, and conclude with a clear resolution or clarification.

3. \textbf{Role Definitions:}
- \texttt{"user"}: Simulate diverse user types (student, researcher, novice).
- \texttt{"assistant"}: Provide accurate, clear, and pedagogical explanations. Avoid jargon or define it clearly.

4. \textbf{Output Format:}
Return a single JSON list where each element represents one conversation—a list of message objects with \texttt{"role"} and \texttt{"content"} keys.
Example:
\texttt{[\{["role": "user", "content": "..."], ["role": "assistant", "content": "..."]\}, ...]}. \\

\end{longtable}
\normalsize

\end{document}